\RequirePackage[svgnames]{xcolor}

\documentclass[11pt,logo,letterpaper]{berkeley}
\usepackage[comma,authoryear,compress]{natbib}
\usepackage[all]{hypcap}
\usepackage{algorithm}
\usepackage{algorithmicx}
\usepackage{algpseudocode}
\usepackage{bigstrut}
\usepackage{mathtools}
\usepackage{mathrsfs}
\DeclareFontFamily{U}{rsfs}{\skewchar\font127}
\DeclareFontShape{U}{rsfs}{m}{n}{%
  <-6> rsfs5
  <6-8> rsfs7
  <8-> rsfs10
}{}
\usepackage{nicefrac}
\usepackage{dsfont}
\expandafter\def\csname ver@subfig.sty\endcsname{}
\usepackage{subcaption}
\usepackage{cleveref}
\usepackage{bxcoloremoji}
\usepackage{wrapfig}
\usepackage{lipsum}
\usepackage{stackengine}
\usepackage{tikz}
\usepackage{rotating}
\usepackage{etoolbox}

\hypersetup{hypertexnames=false}
\newcommand{\benchname}{EEG-Arena}

\makeatletter
\@ifpackagelater{microtype}{2025/02/12}{}{%
  \long\def\MT@skip@showhyphens@check#1[#2]#3{}%
  \long\def\MT@skip@showhyphens@starcheck*#1[#2]#3{}%
  \patchcmd{\MT@setup@expansion}
    {\MT@exp@cs\CheckCommand{showhyphens }}
    {\MT@exp@cs\MT@skip@showhyphens@check{showhyphens }}{}{}%
  \patchcmd{\MT@setup@expansion}
    {\CheckCommand*\showhyphens}
    {\MT@skip@showhyphens@starcheck*\showhyphens}{}{}%
}
\makeatother

\usepackage{multirow}
\usepackage{array}
\usepackage{longtable}
\usepackage{makecell}
\usepackage{adjustbox}
\usepackage{float}
\usepackage{placeins}
\usepackage{pdflscape}

\newcommand{\resultmissing}{%
  \raisebox{-0.25ex}{\char126\relax}%
}

\input{appendix_files/appendix_body_setup.tex}

\tcbuselibrary{most}
\definecolor{rqblue}{HTML}{298CCC}
\definecolor{rqblueDark}{HTML}{1E5F88}
\definecolor{rqblueBg}{HTML}{F2F8FC}
\definecolor{findingred}{HTML}{C84A4A}
\definecolor{findingredBg}{HTML}{FFF4F4}

\newcommand{\evalquestion}[2]{%
  \par
  \begin{tcolorbox}[
    enhanced,
    colback=rqblueBg,
    colframe=rqblue!70!black,
    boxrule=0.65pt,
    arc=2.4mm,
    boxsep=0pt,
    left=13mm,
    right=3.2mm,
    top=2.6mm,
    bottom=2.6mm,
    before skip=7pt plus 2pt minus 1pt,
    after skip=8pt plus 2pt minus 1pt,
    overlay={
      \node[
        circle,
        fill=rqblue,
        text=white,
        font=\sffamily\bfseries\scriptsize,
        minimum size=9.2mm,
        inner sep=0pt
      ] at ([xshift=6.2mm]frame.west) {#1};
    }
  ]
    \raggedright
    {\sffamily\small #2}
  \end{tcolorbox}%
  \par
}

\newcommand{\resultfinding}[2]{%
  \par
  \begin{tcolorbox}[
    enhanced,
    colback=findingredBg,
    colframe=findingred!75!black,
    title={#1},
    coltitle=white,
    fonttitle=\sffamily\bfseries\small,
    attach boxed title to top left={
      xshift=3.2mm,
      yshift*=-\tcboxedtitleheight/2
    },
    boxed title style={
      enhanced,
      colback=findingred,
      colframe=findingred,
      boxrule=0pt,
      arc=1.1mm,
      outer arc=1.1mm,
      boxsep=0pt,
      left=2.4mm,
      right=2.4mm,
      top=0.8mm,
      bottom=0.8mm
    },
    boxrule=0.65pt,
    arc=2.4mm,
    boxsep=0pt,
    left=3.2mm,
    right=3.2mm,
    top=3.3mm,
    bottom=2.4mm,
    before skip=9pt plus 2pt minus 1pt,
    after skip=8pt plus 2pt minus 1pt
  ]
    \raggedright
    {\sffamily\small #2}
  \end{tcolorbox}%
  \par
}

\newcommand{\mainfindinglabel}[1]{%
  \smash{%
    \tikz[baseline=(mainfindingnumber.base)]{%
      \node[
        circle,
        fill=findingred,
        text=white,
        font=\sffamily\bfseries\scriptsize,
        minimum size=4.5mm,
        inner sep=0pt
      ] (mainfindingnumber) {#1};%
    }%
  }%
}

\newtcolorbox{mainfindingsbox}{
  enhanced jigsaw,
  breakable,
  colback=findingredBg!45!white,
  colframe=findingred!55!black,
  boxrule=0.5pt,
  arc=1.6mm,
  boxsep=0pt,
  left=2.6mm,
  right=2.6mm,
  top=2.0mm,
  bottom=2.0mm,
  before skip=5pt plus 1pt minus 1pt,
  after skip=7pt plus 2pt minus 1pt,
  fontupper=\normalfont\normalsize,
  before upper={%
    \begin{enumerate}[
      label=\protect\mainfindinglabel{\arabic*},
      leftmargin=0pt,
      itemindent=6.8mm,
      labelwidth=4.8mm,
      labelsep=1.3mm,
      align=left,
      itemsep=6pt plus 1pt minus 1pt,
      parsep=0pt,
      topsep=0pt,
      partopsep=0pt
    ]%
  },
  after upper={\end{enumerate}}
}

\definecolor{rq1}{HTML}{2563EB}
\definecolor{rq2}{HTML}{0891B2}
\definecolor{rq3}{HTML}{7C3AED}
\definecolor{rq4}{HTML}{EA580C}

\definecolor{rq1Dark}{HTML}{1E40AF}
\definecolor{rq2Dark}{HTML}{155E75}
\definecolor{rq3Dark}{HTML}{5B21B6}
\definecolor{rq4Dark}{HTML}{9A3412}

\newcommand{\x}{\mathbf{x}}

\definecolor{blanchedalmond}{rgb}{1.0, 0.92, 0.8}
\definecolor{carmine}{rgb}{0.59, 0.0, 0.09}
\definecolor{lightblue}{rgb}{0.22,0.45,0.70}%

\renewcommand{\mathbf}{\boldsymbol}

\makeatletter
\def\Ddots{\mathinner{\mkern1mu\raise\p@
\vbox{\kern7\p@\hbox{.}}\mkern2mu
\raise4\p@\hbox{.}\mkern2mu\raise7\p@\hbox{.}\mkern1mu}}
\makeatother

\definecolor{amaranth}{rgb}{0.9, 0.17, 0.31}
\definecolor{antiquebrass}{rgb}{0.8, 0.58, 0.46}
\definecolor{antiquefuchsia}{rgb}{0.57, 0.36, 0.51}
\definecolor{chromeyellow}{rgb}{0.31, 0.47, 0.26}

\newtcolorbox{AIbox}[2][]{aibox,title=#2,#1}
\definecolor{lightblue}{rgb}{0.22,0.45,0.70}%
\definecolor{Gray}{gray}{0.95}
\definecolor{Cornsilk}{rgb}{1.0, 0.97, 0.86}

\title{Benchmarking EEG Foundation Models at Scale: Lessons from 20,000 Evaluations}

\runningtitle{Benchmarking EEG Foundation Models at Scale}

\renewcommand\Affilfont{\centering\normalfont\fontsize{10}{14}\selectfont}

\author{
  Zhige Chen$^{1}$,
  Shu Peng$^{1}$,
  Chengxuan Qin$^{2,3}$,
  Rui Liu$^{1}$,
  Rui Yang$^{3}$,
  Kay Chen Tan$^{1}$,
  Jibin Wu$^{1}$
}

\affil{$^{1}$Department of Data Science and Artificial Intelligence, The Hong Kong Polytechnic University \\
  $^{2}$Department of Computer Science, University of Liverpool \\
  $^{3}$Academy of Artificial Intelligence and Advanced Technology, Xi'an Jiaotong-Liverpool University}

\correspondingauthor{The first three authors contribute equally.}

\begin{document}

\begin{abstract}
Electroencephalography (EEG) foundation models (FMs) promise transferable neural representations, yet their advantages over strong supervised baselines and their prospects for further scaling remain unclear. To address these questions, we introduce \textbf{\benchname{}}, an open-source benchmark covering 30 EEG FMs and 25 supervised baselines evaluated on 57 downstream tasks from 23 public datasets. Through more than 20,000 evaluations across five experimental protocols, we assess downstream performance, pretraining benefits, model size scaling, pretraining data scaling, and robustness to channel configuration. We find that (1) EEG FMs outperform strong task-specific supervised baselines on most evaluated tasks, particularly under non-bipolar settings; (2) compared with architecture-matched supervised training from scratch, pretraining improves both early optimization and final downstream performance, with larger and more consistent gains as more labeled downstream data become available; (3) existing EEG FMs do not exhibit a consistent positive relationship between parameter count and downstream performance; (4) under a fixed architecture, increasing the pretraining data scale yields sustained downstream gains; and (5) channel-flexible FMs achieve higher absolute performance than channel-constrained models across most evaluated channel configurations. Together, these findings demonstrate the downstream value of EEG FMs and identify pretraining data expansion as a promising direction for further progress. To support continued research, we release \benchname{} as an open-source evaluation framework that provides shared infrastructure for reproducible benchmarking, model comparison, and community-driven development.

\vspace{5mm}

\coloremojicode{1F4C5} \textbf{Date}: Sep 26, 2026

\coloremojicode{1F4E7} \textbf{Correspondence}: Jibin Wu~(\href{mailto:jibin.wu@polyu.edu.hk}{jibin.wu@polyu.edu.hk})

\coloremojicode{1F3C6} \textbf{Leaderboard}: \href{https://eeg.fm/leaderboards}{eeg.fm/leaderboards}

\coloremojicode{1F310} \textbf{Platform}: \href{https://eeg.fm/benchmarks}{eeg.fm/benchmarks}

\end{abstract}

\maketitle
\vspace{3mm}

\section{Introduction}
\label{sec:intro}
Electroencephalography (EEG) provides a non-invasive means of monitoring brain activity, supporting applications ranging from clinical assessment to brain--computer interfaces (BCIs). Developing models that generalize across these applications remains challenging because EEG recordings vary substantially across subjects, acquisition systems, electrode configurations, and tasks, while labeled data are often scarce. These challenges have motivated EEG foundation models (FMs), which seek to learn reusable neural representations through large-scale pretraining and transfer them to downstream tasks~\cite{kostas2021bendr,yang2023biot,wang2024eegpt,jiang2024large,wang2025cbramod,wang2025eegmamba}. Recent models have shown promising results, suggesting that pretraining could reduce reliance on learning task-specific representations from scratch. However, evidence from individual models and selected tasks under model-specific experimental settings does not establish whether these advantages generalize across tasks.

\begin{table}[!t]
  \caption{Comparison of representative EEG FM benchmarks. The numbers denote the included EEG FMs, supervised task-specific deep learning (DL) baselines, datasets, and tasks.}
  \label{tab:benchmarks}
  \centering
  \footnotesize
  \setlength{\tabcolsep}{3pt}
  \begin{tabular}{lcccc}
    \toprule
    Benchmark & EEG FMs & Supervised DL baselines & Datasets & Tasks \\
    \midrule
    Lee \emph{et al.}~\cite{lee2025large} & 3 & 2 & 5 & 5 \\
    AdaBrain-Bench~\cite{wu2025adabrainbench} & 4 & 4 & 13 & 13 \\
    EEG-FM-Bench~\cite{xiong2025eegfmbench} & 7 & --- & 14 & 14 \\
    EEG-Bench~\cite{kastrati2025eegbench} & 3 & --- & 14 & 11 \\
    EEG-FM-Compass~\cite{liu2026eegfmbenchmark} & 15 & 7 & 13 & 13 \\
    Brain4FMs~\cite{shen2026brain4fms} & 17 & 5 & 21 & 21 \\
    Yang \emph{et al.}~\cite{yang2026areeeg} & 8 & 4 & 10 & 11 \\
    NeuralBench~\cite{banville2026neuralbench} & 6 & 8 & \textbf{94} & \textbf{97} \\
    NeuroAtlas~\cite{kontras2026neuroatlas} & 10 & 6 & 42 & 61 \\
    OmniEEG-Bench~\cite{lu2026omnieeg} & 10 & 2 & 54 & 58 \\
    OpenEEG-Bench~\cite{guetschel2026toward} & 5 & --- & 10 & 10 \\
    \textbf{\benchname{} (ours)} & \textbf{30} & \textbf{24\textsuperscript{*}} & 23 & 57 \\
    \bottomrule
  \end{tabular}
\vspace{2pt}
\parbox{\linewidth}{%
  \footerfont\itshape
  \textsuperscript{*} We benchmarked 25 baseline models, including 24 DL models and 1 traditional model.
}
\end{table}

Recent benchmarks have begun to address this limitation by expanding model and dataset coverage, as summarized in Table~\ref{tab:benchmarks}. Nevertheless, strong task-specific supervised baselines remain underrepresented, leaving uncertainty about whether EEG FMs improve upon the most competitive alternatives within each paradigm. Differences in input representations, downstream datasets, and adaptation protocols further complicate comparisons across benchmarks. Even when performance gains are observed, their sources remain difficult to identify: comparisons among existing checkpoints may reflect simultaneous differences in architecture, parameter count, and pretraining data. Furthermore, final performance alone also offers limited insight into how pretraining benefits downstream adaptation. A reproducible benchmark is therefore needed to validate the practical value of EEG FMs and clarify the conditions under which their benefits emerge.

To meet this need, we introduce \textbf{\benchname{}}, an open-source benchmark covering 30 EEG FMs and 25 strong supervised baselines across 57 downstream tasks from 23 datasets spanning seven EEG paradigms. Its software framework provides standardized data, model, and evaluation interfaces while preserving model-specific preprocessing and input requirements. Using this infrastructure, we conduct more than 20,000 evaluations through five experimental protocols. These protocols combine broad cross-model comparisons with architecture-matched experiments that isolate the contribution of pretraining, empirical analyses of model size scaling, controlled pretraining data scaling under a fixed architecture, and evaluations of robustness to changes in channel configuration.

By jointly examining these dimensions, \benchname{} moves beyond a conventional performance leaderboard to clarify when EEG pretraining is beneficial and which factors may drive further progress. Our results show that pretrained EEG FMs outperform strong supervised baselines on most evaluated tasks and provide meaningful transfer benefits over supervised training from scratch. However, these benefits cannot be predicted from parameter count alone: existing EEG FMs exhibit no consistent positive relationship between model size and downstream performance. Controlled data scaling experiments instead demonstrate that increasing pretraining data scale under a fixed architecture yields sustained downstream gains. To make these analyses reproducible and extensible, we release an open-source software package and online leaderboard that support the integration of new models, datasets, and evaluation protocols. \benchname{} thus provides both a systematic assessment of current EEG FMs and shared infrastructure for guiding future model development toward the needs of neuroscience and BCI applications.

\begin{figure}[!t]
  \centering
  \includegraphics[width=\linewidth]{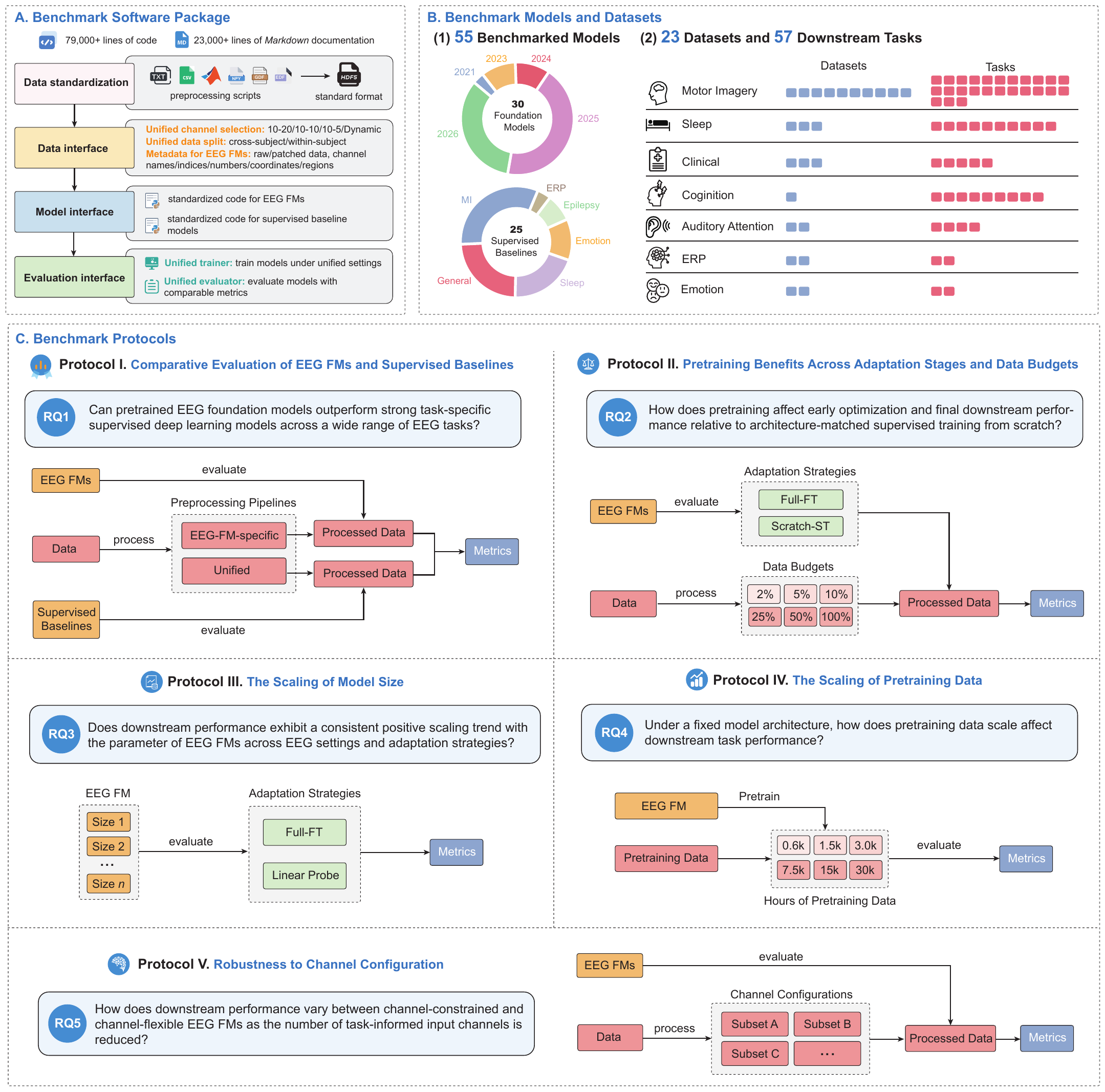}
  \caption{
    Overview of the \benchname{} benchmark framework.
  }\label{fig:eeg2_framework}
\end{figure}

\section{Method}
\label{sec:methods}
To facilitate fair and reproducible benchmarking across EEG FMs, we introduce an open-source software package, illustrated in Fig.~\ref{fig:eeg2_framework}A. The package provides standardized interfaces for data preprocessing, model loading, and evaluation. Fig.~\ref{fig:eeg2_framework}B summarizes the models and datasets covered by \benchname{}, while Fig.~\ref{fig:eeg2_framework}C presents five benchmarking protocols designed to systematically evaluate current EEG FMs. The following sections describe these components in detail.

\subsection{Benchmark Software Package and Evaluation Setup}
\label{sec:benchmark_package}
\paragraph{Data standardization and interface.} The data layer first converts heterogeneous EEG files into HDF5 data format. Then, the data interface provides raw or patched inputs with channel metadata for EEG FMs. Notably, this data interface supports within-subject (WS), cross-subject (CS), and leave-one-subject-out splits, together with 10--20, 10--10, 10--5, and dynamic channel selection.

\paragraph{Model and evaluation interfaces.} The model interface standardizes model initialization, model-specific preprocessing and input construction, and task head configuration for EEG FMs and supervised baselines. The evaluation interface provides unified procedures for optimization, checkpoint selection, metric computation, and result logging while accommodating each model's input requirements.

For each comparison, all supported models are evaluated using identical training, validation, and test splits. The checkpoint achieving the best validation performance is selected, and its performance on the held-out test set is reported. To prevent pretraining data leakage, results are excluded from overall comparisons when a model's pretraining data directly overlap with the corresponding downstream dataset. Model--task combinations that are incompatible because of constraints on channel count or configuration, input length, or input format are marked as unsupported and excluded from ranking analyses.

\subsection{Benchmark Models and Datasets}
\label{sec:benchmark_scope}

\paragraph{Benchmark Models.} Table~\ref{tab:model_baseline_overview_main} summarizes the 30 publicly available EEG FMs and 25 strong supervised baselines evaluated in the \benchname{}. The included datasets and models were selected based on their open-source availability, citation number, and community recognition. Detailed descriptions of the EEG FMs and supervised baselines are provided in Appendix Tables~\ref{tab:fms} and~\ref{tab:baselines}, respectively.

\begin{table}[!htbp]
  \setlength{\abovecaptionskip}{1pt}
  \setlength{\belowcaptionskip}{1pt}
  \caption{EEG FMs and supervised baselines evaluated in \benchname{}.}
  \label{tab:model_baseline_overview_main}
  \centering
  \tiny
  \setlength{\tabcolsep}{2.0pt}
  \renewcommand{\arraystretch}{0.64}
  \resizebox{\linewidth}{!}{%
  \begin{tabular}{llllll@{\hspace{0.8em}}llll}
    \toprule
    \multicolumn{6}{c}{EEG FMs} & \multicolumn{4}{c}{Supervised baselines} \\
    \cmidrule(r){1-6}\cmidrule(l){7-10}
    Model & Year & Variant & Parameters & Architecture\textsuperscript{*} & Channel input strategy & Model & Year & Architecture\textsuperscript{*} & Scope \\
    \midrule
    BENDR~\cite{kostas2021bendr} & 2021 & --- & 157.0M & Transformer & Channel-constrained & EEGNet~\cite{lawhern2018eegnet} & 2018 & CNN & General \\
    BIOT~\cite{yang2023biot} & 2023 & shhs-prest-18ch & 3.3M & Transformer & Channel-constrained & ShallowConvNet~\cite{schirrmeister2017deepconvnet} & 2017 & CNN & General \\
    BIOT~\cite{yang2023biot} & 2023 & prest-16ch & 3.3M & Transformer & Channel-constrained & DeepConvNet~\cite{schirrmeister2017deepconvnet} & 2017 & CNN & General \\
    BIOT~\cite{yang2023biot} & 2023 & six-datasets-18ch & 3.3M & Transformer & Channel-constrained & EEGConformer~\cite{song2022eegconformer} & 2023 & CNN+Transformer & General \\
    EEGPT~\cite{wang2024eegpt} & 2024 & --- & 25.3M & Transformer & Channel-constrained & ATCNet~\cite{altaheri2022atcnet} & 2023 & CNN & MI \\
    LaBraM~\cite{jiang2024large} & 2024 & base & 5.8M & Transformer & Channel-constrained & CTNet~\cite{zhao2024ctnet} & 2024 & CNN+Transformer & MI \\
    NeuroGPT~\cite{cui2024neurogpt} & 2024 & --- & 79.5M & Transformer & Channel-constrained & MSVTNet~\cite{liu2024msvtnet} & 2024 & CNN+ViT & MI \\
    CBraMod~\cite{wang2025cbramod} & 2025 & --- & 4.0M & Transformer & Channel-constrained & EEGInceptionERP~\cite{santamaria2020eeginceptionerp} & 2020 & Inception-CNN & ERP \\
    CSBrain~\cite{zhou2025csbrain} & 2025 & --- & 8.90M & Transformer & Channel-constrained & DBConformer~\cite{wang2025dbconformer} & 2025 & CNN+Transformer & General \\
    EEGMamba~\cite{wang2025eegmamba} & 2025 & --- & 3.3M & Mamba & Channel-constrained & FBCNet~\cite{mane2021fbcnet} & 2021 & CNN & MI \\
    HEAR~\cite{chen2025hear} & 2025 & base & 3.1M & Transformer & Channel-flexible & IFNet~\cite{wang2023ifnet} & 2023 & CNN & MI \\
    HEAR~\cite{chen2025hear} & 2025 & large & 6.1M & Transformer & Channel-flexible & EEGWaveNet~\cite{thuwajit2021eegwavenet} & 2022 & CNN & Epilepsy \\
    LEAD~\cite{wang2025lead} & 2025 & base & 3.41M & Transformer & Channel-flexible & SlimSeiz~\cite{lu2025slimseiz} & 2025 & Mamba & Epilepsy \\
    LUNA~\cite{doner2025luna} & 2025 & base & 7.0M & Transformer & Channel-flexible & MSCFormer~\cite{zhao2025mscformer} & 2025 & CNN+Transformer & MI \\
    LUNA~\cite{doner2025luna} & 2025 & large & 43.0M & Transformer & Channel-flexible & TMSA-Net~\cite{zhao2025tmsanet} & 2025 & CNN & MI \\
    LUNA~\cite{doner2025luna} & 2025 & huge & 311.0M & Transformer & Channel-flexible & ADFCNN~\cite{tao2023adfcnn} & 2024 & CNN & MI \\
    mdJPT~\cite{zhang2025multi} & 2025 & --- & 1.0M & Transformer & Channel-constrained & EEGNeX~\cite{chen2024eegnex} & 2024 & CNN & General \\
    NeuroRVQ~\cite{barmpas2025neurorvq} & 2025 & --- & 5.9M & Transformer & Channel-constrained & BiDANN~\cite{li2018bidann} & 2021 & Adversarial Network & Emotion \\
    REVE~\cite{el2025reve} & 2025 & base & 69.0M & Transformer & Channel-flexible & TSception~\cite{ding2022tsception} & 2023 & Inception & Emotion \\
    REVE~\cite{el2025reve} & 2025 & large & 408.0M & Transformer & Channel-flexible & HSLT~\cite{wang2022hslt} & 2022 & Transformer & Emotion \\
    CodeBrain~\cite{ma2025codebrain} & 2026 & --- & 15.2M & Transformer & Channel-constrained & DeepSleepNet~\cite{supratak2017deepsleepnet} & 2017 & CNN+BiLSTM & Sleep \\
    ST-EEGFormer~\cite{yang2026areeeg} & 2026 & small & 32.7M & Transformer & Channel-constrained & SleepEEGNet~\cite{mousavi2019sleepeegnet} & 2019 & CNN+BiLSTM & Sleep \\
    ST-EEGFormer~\cite{yang2026areeeg} & 2026 & base & 110.9M & Transformer & Channel-constrained & SleepTransformer~\cite{phan2022sleeptransformer} & 2022 & Transformer & Sleep \\
    ST-EEGFormer~\cite{yang2026areeeg} & 2026 & large & 328.4M & Transformer & Channel-constrained & U-Sleep~\cite{perslev2021usleep} & 2021 & U-Net & Sleep \\
    SleepFM~\cite{thapa2026sleepfm} & 2026 & sleeping-staging & 6.12M & Transformer & Channel-flexible & YASA~\cite{vallat2021yasa} & 2021 & Spectral & Sleep \\
    SleepFM~\cite{thapa2026sleepfm} & 2026 & diagnosis & 5.80M & Transformer & Channel-flexible &  &  &  & \\
    SleepFM~\cite{thapa2026sleepfm} & 2026 & upstream & 4.83M & Transformer & Channel-flexible &  &  &  & \\
    SleepGPT~\cite{huang2026sleepgpt} & 2026 & large & 134M & Transformer & Channel-constrained &  &  &  & \\
    TFM~\cite{pradeepkumar2026tfm} & 2026 & CHBMIT & 1.89M & Transformer & Channel-constrained &  &  &  & \\
    Uni-DMFM~\cite{chen2026unidmfm} & 2026 & base & 29.42M & Transformer & Channel-flexible &  &  &  & \\
    \bottomrule
  \end{tabular}}
\vspace{2pt}
\parbox{\linewidth}{%
  \footerfont\itshape
  \textsuperscript{*} Architecture refers to the primary backbone of each model.
}
\end{table}

\paragraph{Evaluated Datasets and Tasks.} \benchname{} covers 23 datasets and 57 downstream tasks across 7 paradigms, including motor imagery (MI), sleep detection (Sleep), cognitive and workload detection (Cognitive), clinical-related tasks (Clinical), auditory attention decoding (AAD), event-related potential (ERP), and emotion recognition (Emotion) tasks. The detailed dataset information and the definition of each task are provided in Appendix~\ref{app:datasets}. For tasks with reliable subject metadata, we report both WS and CS results.

\subsection{Benchmark Protocol I: Comparative Evaluation of EEG FMs and Supervised Baselines}
\label{sec:protocol_overall}

\evalquestion{RQ1}{Can pretrained EEG FMs outperform strong task-specific supervised baseline models across a wide range of EEG tasks?}

\paragraph{Benchmark Metrics.}
To facilitate performance comparisons across models, we introduce the Average Rank and Top-$5$ Count as metrics to characterize overall performance and ranking consistency across models and downstream tasks. Rankings are computed separately for bipolar and non-bipolar EEG settings. 

\textit{(1) Average Rank.}
Let $\mathcal{T}_m$ denote the set of valid downstream tasks for model $m$. For each task $t$, let
$r_{m,t}^{\mathrm{BA}}$,
$r_{m,t}^{\mathrm{F1}}$, and
$r_{m,t}^{\kappa}$
denote the ranks of model $m$ according to balanced accuracy, macro-F1 score, and Cohen's $\kappa$, respectively. We compute the average rank of model $m$ as

\begin{equation}
\bar{r}_m
=
\frac{1}{|\mathcal{T}_m|}
\sum_{t\in\mathcal{T}_m}
\operatorname{rank}_{t}
\left(
\frac{
r_{m,t}^{\mathrm{BA}}
+
r_{m,t}^{\mathrm{F1}}
+
r_{m,t}^{\kappa}
}{3}
\right),
\label{eq:metric_mean_rank}
\end{equation}

where $\operatorname{rank}_{t}(\cdot)$ denotes the rank among all models with valid results on task $t$, based on the mean of the three metric-specific ranks. A lower $\bar{r}_m$ indicates better overall ranking performance. 

\textit{(2) Top-$5$ Count.}
The top-$5$ count of model $m$ is defined as the number of valid tasks on which it ranks among the five best-performing models:

\begin{equation}
C_5(m)
=
\sum_{t\in\mathcal{T}_m}
\mathbb{I}\!\left(r_{m,t}\leq 5\right),
\label{eq:metric_topk}
\end{equation}

where $\mathbb{I}(\cdot)$ is the indicator function. A larger $C_5(m)$ indicates that the model more frequently achieves top ranked performance across downstream tasks.

\subsection{Benchmark Protocol II: Pretraining Benefits Across Adaptation Stages and Data Budgets}
\label{sec:protocol_pretraining}

\evalquestion{RQ2}{
How does pretraining improve early optimization and final downstream performance relative to supervised training from scratch, and how does this benefit vary with the amount of labeled downstream data?
}

\paragraph{Protocol Design.}
We compare full fine-tuning (Full-FT) with architecture-matched scratch supervised training (Scratch-ST). In Full-FT, the backbone is initialized with the model's pretrained weights, and both the backbone and the task-specific prediction head are updated using labeled downstream data. In contrast, Scratch-ST uses the same backbone architecture and prediction head but initializes both components randomly, without using pretrained weights. This paired design isolates the contribution of pretraining while controlling for model architecture. To examine how the pretraining benefit depends on the amount of downstream supervision, we conduct these comparisons under six training data budgets: $\{2\%,5\%,10\%,25\%,50\%,100\%\}$. The validation and test sets remain fixed across all conditions, while the training set is subsampled according to the specified budget.

\paragraph{Benchmark Metric.}
Let $\alpha\in\{\mathrm{Full\text{-}FT},\mathrm{Scratch\text{-}ST}\}$ denote the training strategy, and let $d$ denote the training data budget. We use $s_{m,t}(d,\alpha)$ to denote the balanced accuracy for binary classification or the macro-F1 score for multiclass classification, obtained by model architecture $m$ on task $t$ under data budget $d$ and strategy $\alpha$. Let $\mathcal{T}_m$ denote the set of downstream tasks with results for both strategies. The average performance of model $m$ is defined as
\begin{equation}
\bar{s}_{m}(d,\alpha)
=
\frac{1}{|\mathcal{T}_m|}
\sum_{t\in\mathcal{T}_m}
s_{m,t}(d,\alpha).
\label{eq:metric_mean_score}
\end{equation}

We define the average pretraining gain as
\begin{equation}
\bar{G}_{m}(d)
=
\bar{s}_{m}\!\left(d,\mathrm{Full\text{-}FT}\right)
-
\bar{s}_{m}\!\left(d,\mathrm{Scratch\text{-}ST}\right).
\label{eq:metric_pretraining_gain}
\end{equation}
A positive value of $\bar{G}_{m}(d)$ indicates that Full-FT outperforms its
architecture-matched Scratch-ST counterpart on average across
the evaluated tasks under the data budget $d$.

\subsection{Benchmark Protocol III: The Scaling of Model Size}

\evalquestion{RQ3}{
Does downstream performance exhibit a consistent positive scaling trend with the parameter count of EEG FMs across EEG settings and adaptation strategies?
}

\paragraph{Protocol Design.}
We compare EEG FMs and their available model variants spanning different parameter scales. Each model is evaluated under Full-FT and linear probing (Linear Probe) separately on bipolar and non-bipolar EEG settings, using the same data splits and evaluation protocols. In Linear Probe, the backbone is frozen and only the prediction head is trained. Thus, Full-FT evaluates performance after end-to-end adaptation, while Linear Probe provides a more direct assessment of the quality of the pretrained representations with minimal task-specific adaptation. 

For each model variant, we compute its average performance across valid downstream tasks and examine the relationship between parameter count and downstream performance. When multiple variants are available within the same model family, we further compare their performance across model scales. It is worth mentioning that because EEG FMs differ in architecture, pretraining data scale and quality, training strategy, and optimization procedure, this analysis characterizes the empirical relationship between model scale and downstream performance rather than the isolated causal effect of parameter count. A controlled scaling study that holds these factors fixed would be required to establish such a causal relationship.

\paragraph{Benchmark Metrics.} We use the same task-specific evaluation metrics defined in Benchmark Protocol II and conduct the analysis separately for each adaptation strategy and EEG setting. We then analyze the relationship between average downstream performance and parameter count, both across EEG FM families and within model families containing multiple scale variants.

\subsection{Benchmark Protocol IV: The Scaling of Pretraining Data}
\label{sec:protocol_pretraining_data} 

\evalquestion{RQ4}{
Under a fixed model architecture, how does pretraining data scale affect downstream task performance?
}

\paragraph{Protocol Design.}  Establishing how pretraining data scale affects downstream performance is essential for guiding future efforts to scale EEG FM pretraining. To isolate the effect of pretraining data scale, we conduct a controlled experiment using a fixed model architecture. Specifically, as shown in Figure~\ref{fig:eeg2_framework}C.IV, we select \textit{REVE-base}, the top-ranked EEG FM under the non-bipolar setting in Protocol I, as the backbone and pretrain it on six data scales: 600, 1,500, 3,000, 7,500, 15,000, and 30,000 hours. The model architecture, parameter count, pretraining objective, and optimization algorithm are held constant across all six experiments, with only the amount of pretraining data varied. Each pretrained checkpoint is subsequently evaluated using Full-FT and Linear Probe under both within-subject and cross-subject settings, following evaluation procedures described in Protocol I.

\paragraph{Benchmark Metrics.} We evaluate the six pretrained checkpoints using average performance, with results reported separately for the within-subject and cross-subject settings. Within each setting, average performance is computed by assigning equal weight to each task. To further characterize how pretraining data scale affects downstream adaptation, we quantify the parameter and representational changes induced by Full-FT using the following metrics.

\textit{(1) Relative Parameter Displacement.}
To quantify the extent to which the pretrained model parameters are modified during Full-FT, we define the relative parameter displacement as

\begin{equation}
P_{s,t}
=
\frac{
\left\|
\boldsymbol{\theta}_{s,t}^{\mathrm{post}}
-
\boldsymbol{\theta}_{s}^{\mathrm{pre}}
\right\|_2
}{
\left\|
\boldsymbol{\theta}_{s}^{\mathrm{pre}}
\right\|_2
},
\label{eq:metric_parameter_displacement}
\end{equation}
where $s$ denotes the pretraining data scale in hours, and $t$ denotes a downstream task. The terms $\boldsymbol{\theta}_{s}^{\mathrm{pre}}$ and $\boldsymbol{\theta}_{s,t}^{\mathrm{post}}$ denote the model parameters before and after Full-FT on task $t$, respectively. A lower value indicates that less parameter adjustment is required for downstream adaptation.

\textit{(2) Representational Change.}
To quantify how the model representations change during Full-FT, we adopted a centered-kernel-alignment (CKA)-based method~\cite{kornblith2019cka} as 
\begin{equation}
D_{s,t,\ell}
=
1-
\operatorname{CKA}
\left(
\mathbf{Z}_{s,t,\ell}^{\mathrm{pre}},
\mathbf{Z}_{s,t,\ell}^{\mathrm{post}}
\right),
\label{eq:metric_representational_change}
\end{equation}

where $\mathbf{Z}_{s,t,\ell}^{\mathrm{pre}}$ and
$\mathbf{Z}_{s,t,\ell}^{\mathrm{post}}$ denote the network representations of layer $l$ before and after Full-FT, respectively. A larger value of $D_{s,t,\ell}$ indicates a greater change in the learned representations during downstream adaptation.

\subsection{Benchmark Protocol V: Robustness to Channel Configuration}
\label{sec:protocol_montage}

\evalquestion{RQ5}{
How does downstream performance vary between channel-constrained and channel-flexible EEG FMs as the number of task-informed input channels is reduced?
}

\paragraph{Protocol Design.}
EEG acquisition systems often differ in electrode coverage, channel count, and spatial layout because of hardware constraints, recording protocols, and application requirements. These variations can affect both the information available to a model and the compatibility of its input interface, making channel robustness important for reduced-channel, wearable, and resource-constrained applications. 

As summarized in Table~\ref{tab:fms}, EEG FMs are categorized according to their channel-input strategies as either channel-constrained or channel-flexible. Channel-constrained models assume a fixed channel set or a predefined electrode mapping, whereas channel-flexible models can accommodate heterogeneous channel subsets and spatial layouts, often through explicit electrode or positional encodings. To address RQ5, we compare the downstream performance of these two model categories as the number of task-informed input channels is reduced. Specifically, we evaluate ten representative EEG FMs under their native full-channel configurations and a series of compatible, task-informed reduced-channel configurations. The evaluated models comprise six channel-constrained models (i.e., \textit{CBraMod}, \textit{CodeBrain}, \textit{CSBrain}, \textit{EEGPT}, \textit{NeuroRVQ}, and \textit{ST-EEGFormer-small}), and four channel-flexible models (i.e.\textit{REVE-base}, \textit{HEAR-base}, \textit{HEAR-large}, and \textit{LEAD}).

For each dataset--task pair, we construct predefined electrode subsets with varying channel counts. Smaller subsets retain electrodes from task-relevant scalp regions, whereas larger subsets progressively broaden spatial coverage. For example, the motor imagery and cursor control configurations prioritize electrodes over central sensorimotor regions, whereas the P300 target detection and feedback classification configurations prioritize centro-parietal regions. Each model is evaluated only under channel configurations supported by its input interface. The exact channel configurations used for each dataset, task, and model are detailed in Appendix Tables~\ref{tab:channel_subset_inventory} and~\ref{tab:model_specific_channel_inputs}.

\begin{figure}[htbp]
  \centering
  \makebox[\linewidth][c]{\includegraphics[width=1\linewidth]{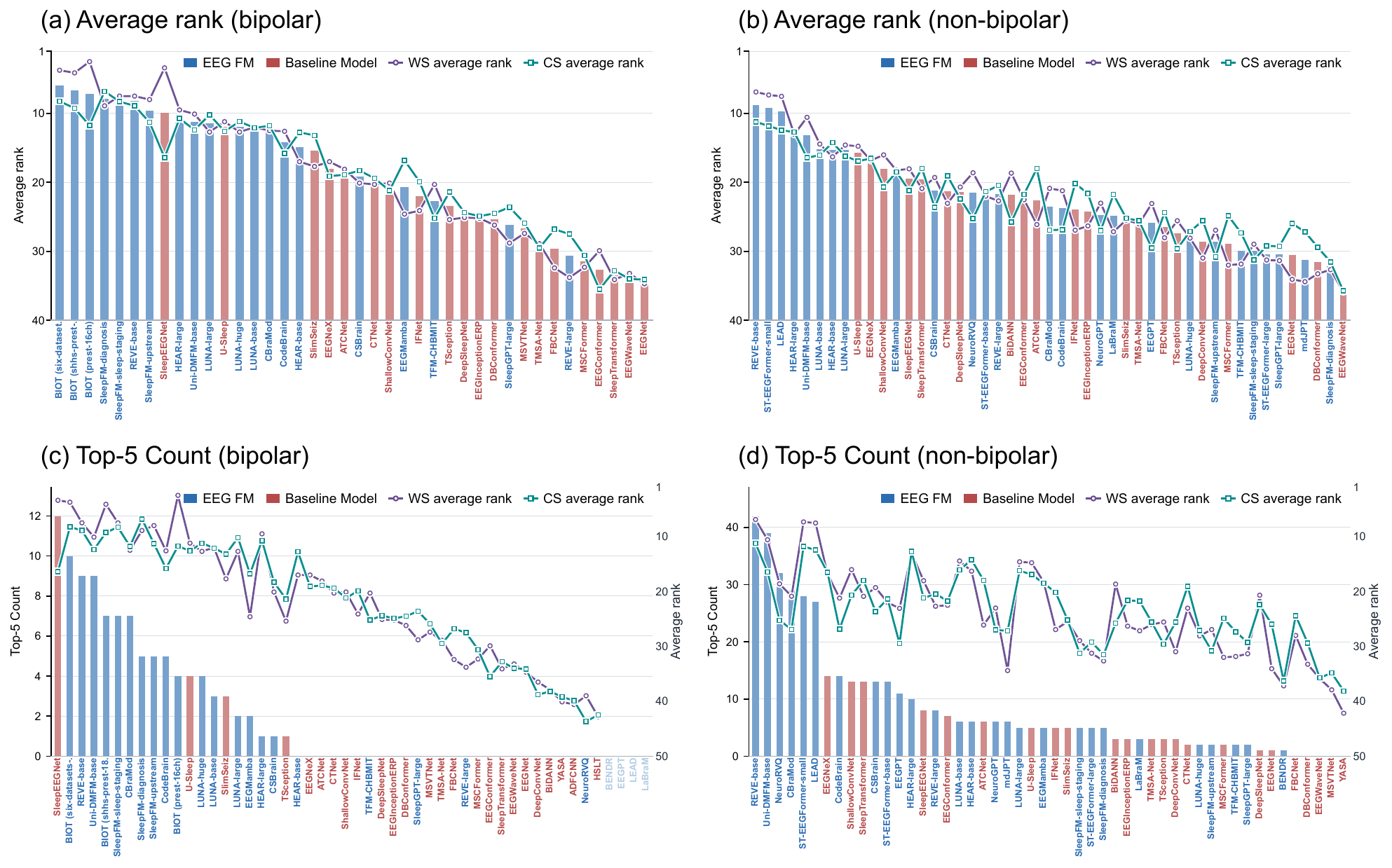}}
  \caption{
    Average rank and top-$5$ count of EEG FMs and supervised baselines under bipolar and non-bipolar EEG settings.
    }
  \label{fig:overall_results_rankings_scaling}
\end{figure}

\paragraph{Benchmark Metrics.}
We evaluate each model under different channel configurations using the task-specific metrics defined in Benchmark Protocol~I. For model $m$ and task $t$, let $c_{\mathrm{f}}$ denote the corresponding full-channel configuration and let $c_{\mathrm{r}}$ denote a reduced-channel configuration. We define the retained-channel ratio as

\begin{equation}
\rho_{m,t}(c_{\mathrm{r}})
=
\frac{
N_{m,t}(c_{\mathrm{r}})
}{
N_{m,t}(c_{\mathrm{f}})
}
\times 100\%,
\label{eq:metric_retained_channel_ratio}
\end{equation}

where $N_{m,t}(c_{\mathrm{r}})$ denotes the number of selected input channels under the reduced-channel configuration $c_{\mathrm{r}}$, and $N_{m,t}(c_{\mathrm{f}})$ denotes the number of channels under the corresponding full-channel configuration $c_{\mathrm{f}}$. For each model and retained-channel ratio, we report the average performance across all valid downstream tasks. This analysis characterizes how downstream performance changes as channel coverage decreases from the full-channel configuration to smaller, task-informed electrode subsets.

\begin{figure}[!htbp]
  \centering
  \makebox[\linewidth][c]{\includegraphics[width=1.0\linewidth]{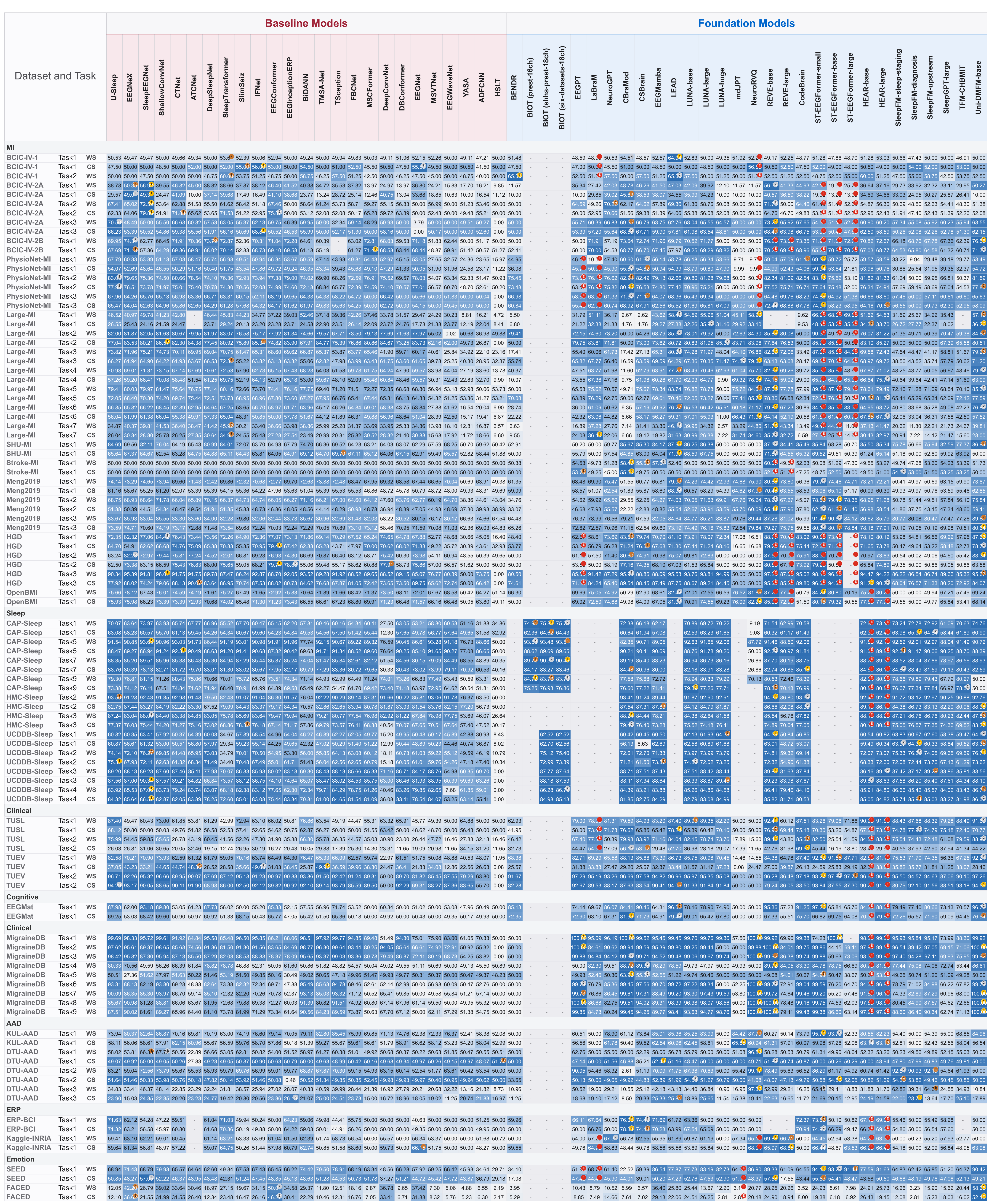}}
  \caption{
    Summary of evaluation results across all models and downstream tasks. For each task, the three best-performing models are indicated by numeric annotations. Models that do not support a particular task are denoted by ``--'', while red ``E'' badges denote tasks whose data overlap with the corresponding pretraining data; these cases are excluded from the ranking analysis. }
  \label{fig:overall_performance_matrix}
\end{figure}

\section{Experimental Results}
\label{sec:results}

This section presents the benchmarking results and addresses the corresponding research question for each evaluation protocol.

\subsection{RQ1: Comparative Evaluation of EEG FMs and Supervised Baselines}
\label{sec:results_overall}

\resultfinding{Main Finding}{
EEG FMs outperform strong supervised baselines on most evaluated tasks, with the largest advantages observed under non-bipolar channel configurations.}

We rank the models using their average rank $\bar{r}$ and top-$5$ count, as shown in Figure~\ref{fig:overall_results_rankings_scaling}. To account for differences in input compatibility, we report rankings separately for bipolar and non-bipolar EEG configurations. The detailed results across all models and datasets are summarized in Figure~\ref{fig:overall_performance_matrix}.

Under non-bipolar configurations, several EEG FMs consistently outperform the supervised baselines. In particular, \textit{REVE-base}, \textit{ST-EEGFormer-small}, \textit{LEAD}, and \textit{HEAR-large} achieve relatively low average ranks and high top-$5$ counts, as shown in Figures~\ref{fig:overall_results_rankings_scaling}b and~\ref{fig:overall_results_rankings_scaling}d. Under bipolar configurations, EEG FMs remain competitive with, and in most cases outperform, the supervised baselines. The \textit{BIOT} variants achieve the lowest average ranks, while \textit{REVE-base} and the \textit{SleepFM} variants also show strong performance in terms of both average rank and top-$5$ count, as shown in Figures~\ref{fig:overall_results_rankings_scaling}a and~\ref{fig:overall_results_rankings_scaling}c. \textit{SleepEEGNet} achieves the highest top-$5$ count; however, this result is partly attributable to this model being evaluated on a larger number of bipolar tasks than the EEG FMs.

Overall, EEG FMs exhibit a clear advantage over supervised baselines in terms of average rank across both bipolar and non-bipolar configurations for CS and WS tasks. These comparisons should nevertheless be interpreted in the context of input compatibility: directly comparing models designed specifically for bipolar inputs, such as \textit{BIOT}, with models developed for non-bipolar inputs may create an unfavorable evaluation setting and underestimate the performance of the former.

\subsection{RQ2: Pretraining Benefits Across Adaptation Stages and Data Budgets}
\label{sec:results_pretraining_effectiveness}

\resultfinding{Main Finding}{
Pretraining provides an early optimization advantage and a sustained downstream performance benefit, with the magnitude and consistency of this benefit generally increasing with the amount of labeled data.
}

\begin{figure}[ht]
  \centering
  \includegraphics[width=\linewidth]{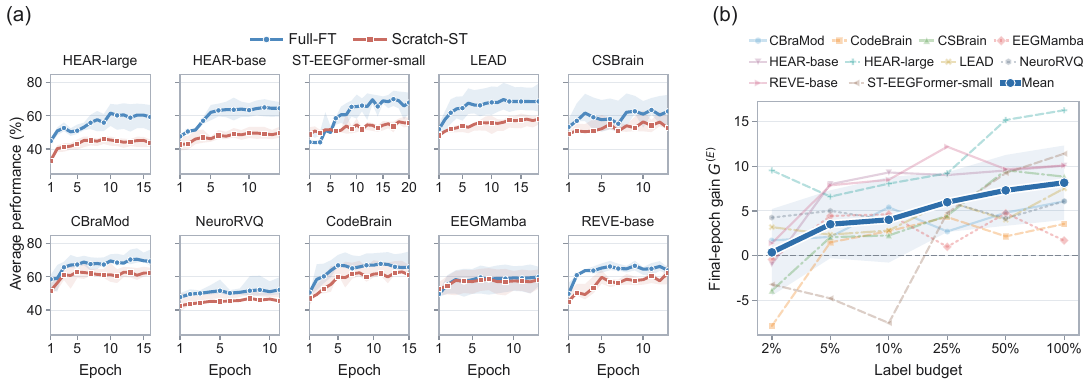}
  \caption{Pretraining benefit analysis across EEG FMs. (a) Learning curves for Full-FT and Scratch-ST, showing that pretraining provides a favorable initialization and sustains its performance advantage throughout fine-tuning. (b) Final-epoch gains increase with the amount of labeled downstream data, suggesting that pretraining enhances the models' ability to exploit additional labeled data. Shaded regions indicate the standard deviation across all the evaluated models.
     }
  \label{fig:pretraining_benefit_eeg_fms}
\end{figure}

We first quantify the benefit of pretraining by comparing Full-FT with architecture-matched Scratch-ST. As shown in Figure~\ref{fig:pretraining_benefit_eeg_fms}(a), eight of the ten models exhibit a positive first-epoch gain,  indicating that pretraining improves optimization from the beginning of downstream adaptation. By the final epoch, all ten models exhibit positive gains, demonstrating that pretraining provides a sustained downstream advantage.

We next examine how the final-epoch gain varies with the downstream training data budget $d$. As shown in Figure~\ref{fig:pretraining_benefit_eeg_fms}(b), the mean final-epoch gain across evaluated models and tasks increases from $0.4\%$ to $8.2\%$ as $d$ grows from $2\%$ to $100\%$, and all ten models exhibit positive gains when $d\geq25\%$. These results suggest that pretraining learns general purpose EEG representations that improve early optimization but require sufficient downstream supervision to be effectively aligned with task-specific objectives and translated into consistent performance gains. 

Overall, pretraining not only provides a favorable initialization but also improves the model's ability to exploit additional labeled data. Although larger labeled datasets generally yield more pronounced gains, the observed early stage improvements demonstrate that pretrained EEG FMs can outperform supervised deep learning models even when downstream labels are limited.

\subsection{RQ3: The Scaling of Model Size}
\label{sec:results_model_size}
\resultfinding{Main Finding}{
Current EEG FMs do not exhibit a consistent positive relationship between model size and downstream performance.
}

\begin{figure}[!ht]
  \centering
  \makebox[\linewidth][c]{\includegraphics[width=0.95\linewidth]{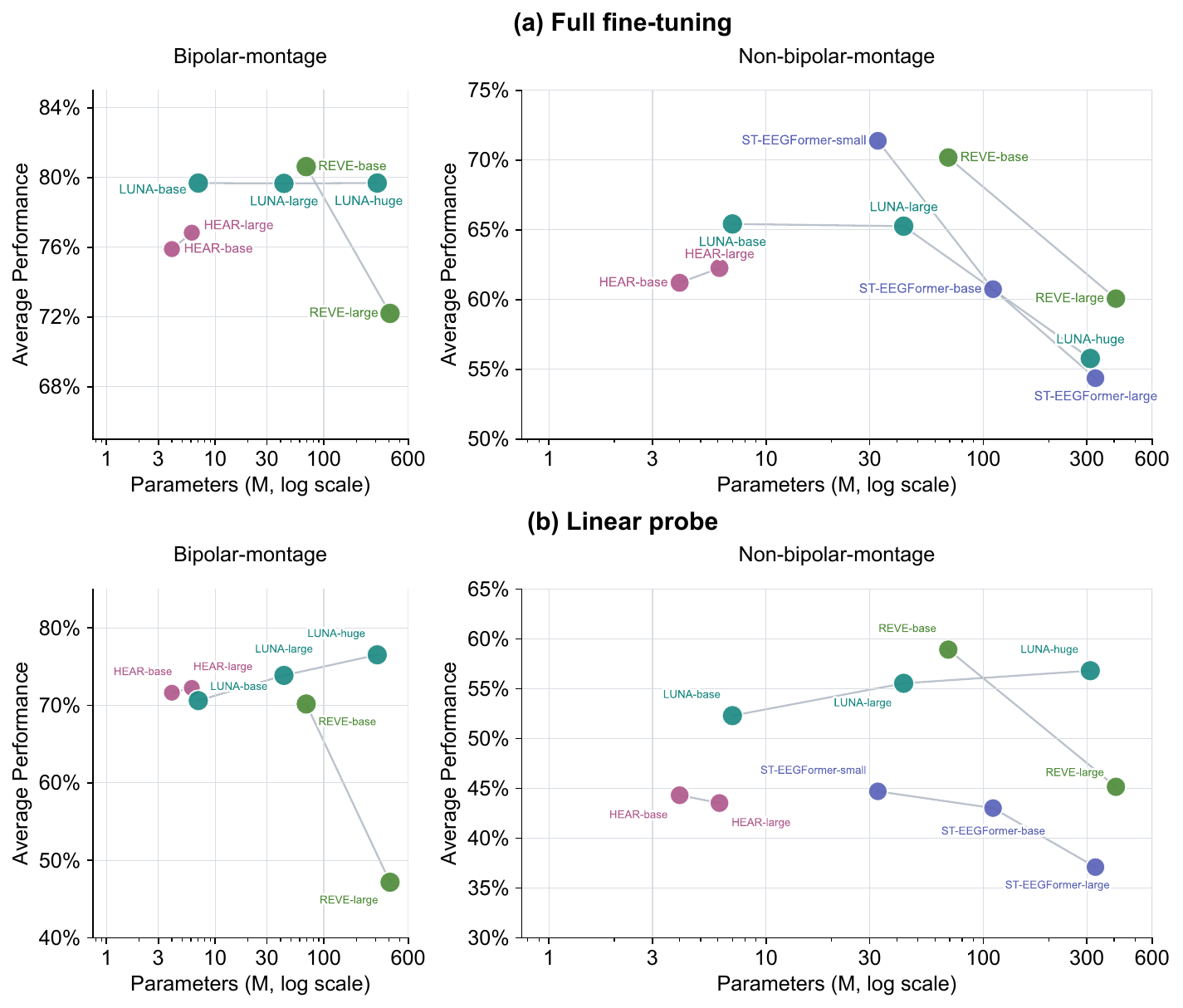}}
  \caption{
    Model size scaling analysis across EEG FMs under (a) Full fine-tuning, and (b) Linear probe. 
  }
  \label{fig:scaling_analysis}
\end{figure}

As shown in Figure~\ref{fig:scaling_analysis}, we examine the empirical relationship between model size, measured by parameter count, and the average downstream performance of EEG FMs under bipolar and non-bipolar settings. Performance does not increase monotonically with model size: larger models do not consistently outperform their smaller counterparts, and performance sometimes declines as the parameter count increases. For example, under bipolar settings, the smaller \textit{REVE-base} achieves higher average performance across tasks than \textit{REVE-large}. Under non-bipolar settings, \textit{REVE-base} and \textit{ST-EEGFormer-small} likewise outperform their larger variants.

Overall, these results provide no evidence of a consistent positive scaling trend among the evaluated EEG FMs. However, because the evaluated model checkpoints differ in architecture, pretraining data scale and quality, and training strategy, the observed differences cannot be attributed to parameter count alone. Controlled experiments that vary model size while holding these factors fixed are needed to characterize the scaling properties of EEG FMs more rigorously.

\subsection{RQ4: The Scaling of Pretraining Data}
\label{sec:results_data_scaling}

\resultfinding{Main Finding}{
With the model architecture and training setup held fixed, increasing the scale of pretraining data consistently improves downstream performance and leads to more parameter-efficient yet broader representational adaptation.
}

\begin{figure}[!htbp]
  \centering
  \includegraphics[width=\linewidth]{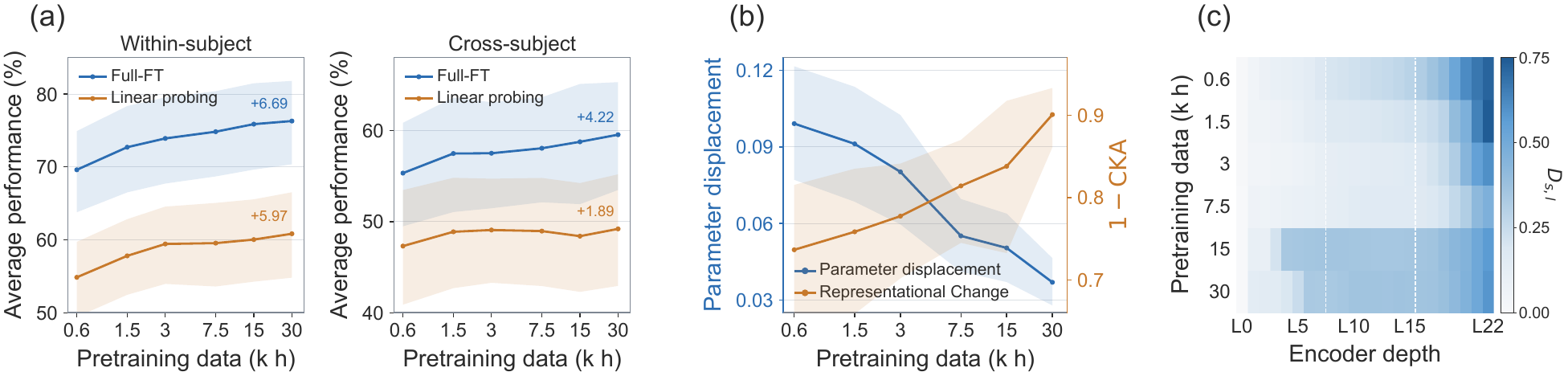}
  \caption{(a) Downstream performance under within-subject and cross-subject settings for Full-FT and Linear Probe across different pretraining data scales. (b) Relative parameter displacement and model output representational change induced by Full-FT. (c) Layer-wise representational change across layers and pretraining data scales.}
  \label{fig:protocol_iii_scaling}
\end{figure}

We first examine how downstream performance varies as the pretraining data scale increases from 600 to 30,000 hours. Figure~\ref{fig:protocol_iii_scaling}(a) summarizes the results under the within-subject and cross-subject settings. Increasing the pretraining data scale consistently improves Full-FT performance, increasing average performance by 6.69 and 4.22 percentage points under the within-subject and cross-subject settings, respectively. To further characterize how pretraining data scale influences downstream adaptation, we analyze the relative parameter displacement and representational changes induced by Full-FT. As shown in Figure~\ref{fig:protocol_iii_scaling}(b), larger-scale pretraining is associated with smaller relative parameter displacement but greater representational change at the model output. Specifically, as the pretraining data scale increases from 600 to 30,000 hours, relative parameter displacement progressively decreases, whereas output-level representational change increases. These results indicate that models pretrained on more data achieve better downstream performance through greater functional adaptation of their learned representations while requiring comparatively smaller changes to the pretrained parameters.

We next investigate where these representational changes occur within the model and how their layerwise distribution varies with pretraining data scale. As shown in Figure~\ref{fig:protocol_iii_scaling}(c), increasing the pretraining data scale broadens representational adaptation from the final layers to the intermediate layers, while the early-layer representations remain relatively stable. At smaller pretraining data scales, adaptation is concentrated primarily in the final layers. Once the pretraining data scale reaches 15,000--30,000 hours, representational changes become more pronounced in the intermediate layers but remain limited in the early layers. 

Taken together, these controlled experiments demonstrate sustained downstream benefits from scaling up pretraining data. Larger-scale pretraining appears to produce representations that can be more effectively reorganized for downstream tasks through relatively modest parameter updates, particularly within the intermediate and final layers. These findings motivate continued efforts to curate and integrate large-scale EEG datasets to advance the development of EEG FMs.

\subsection{RQ5: Robustness to Channel Configuration}
\label{sec:results_channel_configuration}

\resultfinding{Main Finding}{
Channel-flexible EEG FMs achieve higher absolute downstream performance than channel-constrained models across most task-informed channel configurations.
}

\begin{figure}[htbp]
  \centering
  \includegraphics[width=0.5\linewidth]{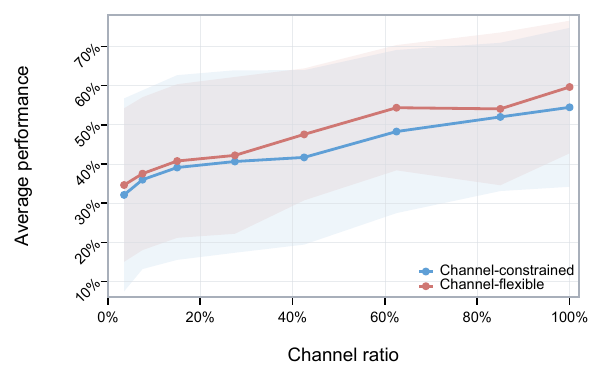}
  \caption{Average downstream performance as a function of the retained channel ratio. Shaded regions indicate the standard deviation across all the evaluated models.
  }
  \label{fig:channel_reduction_experiments}
\end{figure}

Figure~\ref{fig:channel_reduction_experiments} summarizes the average downstream performance of channel-constrained and channel-flexible EEG FMs across different retained channel ratios. Channel-flexible models generally achieve higher absolute performance across all channel reduced settings. This advantage may arise from their ability to accommodate heterogeneous channel subsets and spatial layouts, often through explicit electrode or positional encodings, thereby reducing the input mismatch caused by changes in the available electrodes. In contrast, channel-constrained models rely on fixed channel sets or predefined electrode mappings and may therefore be more sensitive to deviations from their native input configurations. The consistent performance advantage of channel-flexible models across a broad range of retained-channel ratios suggests that flexible channel-input strategies can facilitate deployment across EEG systems with different channel counts and acquisition configurations.

\section{Towards Systematic Evaluation of EEG FMs}
\label{sec:model_profiles}

As EEG FMs continue to evolve, their practical utility cannot be characterized by downstream accuracy alone. A model may achieve strong task performance while requiring a large number of channels, substantial labeled data, or considerable computational resources. Conversely, a smaller or more channel-flexible model may be preferable in a resource-constrained deployment. We therefore advocate a multidimensional evaluation framework that considers task performance, pretraining effectiveness, channel compatibility, and computational efficiency jointly.

Based on the benchmark protocols and results, we define eight evaluation dimensions organized into three categories, as summarized in Table~\ref{tab:model_profile_metrics}: \textit{Task Performance}, \textit{Pre-training Effectiveness}, and \textit{Coverage and Efficiency}. The first category measures performance across bipolar and non-bipolar EEG tasks. The second characterizes the benefits of pretraining during downstream adaptation, including first-epoch and final-epoch gains as well as performance under limited labeled-data budgets. The third captures the practical applicability of each model through channel configuration coverage, parameter efficiency, and inference throughput.

\begin{table}[htbp]
  \centering
  \caption{Definitions and metrics across the eight evaluation dimensions.}
  \label{tab:model_profile_metrics}
  \footnotesize
  \definecolor{profilePerformance}{HTML}{315E91}
  \definecolor{profilePretraining}{HTML}{00866C}
  \definecolor{profileEfficiency}{HTML}{7B4FB8}
  \definecolor{profileAxisOne}{HTML}{5D8CC1}
  \definecolor{profileAxisTwo}{HTML}{709BC7}
  \definecolor{profileAxisThree}{HTML}{35A889}
  \definecolor{profileAxisFour}{HTML}{68BBA4}
  \definecolor{profileAxisFive}{HTML}{91CDBA}
  \definecolor{profileAxisSix}{HTML}{AB8DCE}
  \definecolor{profileAxisSeven}{HTML}{8A62C2}
  \definecolor{profileAxisEight}{HTML}{7B4FB8}
  \newcommand{\profiledimension}[4]{%
    \hangindent=15pt\hangafter=1\relax
    \tikz[baseline=(badge.base)]\node[rectangle,fill=#1,text=#2,inner sep=0pt,minimum width=11pt,minimum height=11pt,font=\sffamily\bfseries\scriptsize](badge){#3};\hspace{4pt}#4}
  \setlength{\tabcolsep}{4pt}
  \renewcommand{\arraystretch}{1.25}
  \begin{tabularx}{\linewidth}{@{}>{\raggedright\arraybackslash}p{0.13\linewidth} >{\raggedright\arraybackslash}p{0.215\linewidth} >{\raggedright\arraybackslash}X >{\raggedright\arraybackslash}p{0.265\linewidth}@{}}
    \toprule
    \textbf{Category} & \textbf{Dimension} & \textbf{Definition} & \textbf{Metric} \\
    \midrule
    \multirow[t]{2}{=}{\textcolor{profilePerformance}{\textbf{Task Performance}}} & \profiledimension{profileAxisOne}{white}{1}{Average rank across bipolar tasks} & Average rank across bipolar EEG tasks, as defined in Protocol~I (Equation~\ref{eq:metric_mean_rank}). & $\bar r_m$ on bipolar tasks \\
    \addlinespace[5pt]
     & \profiledimension{profileAxisTwo}{white}{2}{Average rank across non-bipolar tasks} & Average rank across non-bipolar EEG tasks, as defined in Protocol~I (Equation~\ref{eq:metric_mean_rank}). & $\bar r_m$ on non-bipolar tasks \\
    \midrule
    \multirow[t]{3}{=}{\textcolor{profilePretraining}{\textbf{Pre-training Effectiveness}}} & \profiledimension{profileAxisThree}{white}{3}{Final-epoch\newline pretraining gain} & Final-epoch pretraining gain at the 100\% training-data budget, as defined in Protocol~II (Equation~\ref{eq:metric_pretraining_gain}). &  $\bar{G}_{m}(100\%)$ \\
    \addlinespace[5pt]
     & \profiledimension{profileAxisFour}{white}{4}{First-epoch\newline pretraining gain} & First-epoch pretraining gain at the 100\% training-data budget, as defined in Protocol~II (Equation~\ref{eq:metric_pretraining_gain}). & $\bar{G}_{m}(100\%)$ \\
    \addlinespace[5pt]
     & \profiledimension{profileAxisFive}{white}{5}{Limited labeled-data performance} & Average performance (defined in Protocol~II, Equation~\ref{eq:metric_mean_score}) of Full-FT at 2\%, 5\%, and 10\% training data budgets. & $\displaystyle \frac{1}{3}\sum_{d\in\{2\%,5\%,10\%\}}\bar{s}_m(d,\mathrm{Full\text{-}FT})$ \\
    \midrule
    \multirow[t]{3}{=}{\textcolor{profileEfficiency}{\textbf{Coverage and Efficiency}}} & \profiledimension{profileAxisSix}{white}{6}{Channel configuration coverage\textsuperscript{*}} & Average benchmark-task coverage across bipolar and non-bipolar settings. & --- \\
    \addlinespace[5pt]
     & \profiledimension{profileAxisSeven}{white}{7}{Parameter efficiency} & Reciprocal of the product of the average rank $\bar r_m$ defined in Protocol~I (Equation~\ref{eq:metric_mean_rank}) and model size $P_m$ (millions). & $\displaystyle \frac{1}{\bar r_m P_m}$ \\
    \addlinespace[5pt]
     & \profiledimension{profileAxisEight}{white}{8}{Inference throughput} & Reciprocal of the average task-normalized runtime $\tau_m$ across tasks. & $\displaystyle \frac{1}{\tau_m}$ \\
    \bottomrule
  \end{tabularx}
\vspace{2pt}
\parbox{\linewidth}{%
  \footerfont\itshape
  \textsuperscript{*} Channel configuration refers to the input channel count, and channel reference scheme, including bipolar and non-bipolar settings.
}
\end{table}

These dimensions are complementary and may be prioritized differently across applications. For example, a clinical system with limited annotations may assign greater weight to low-data performance and final-epoch pretraining gain, whereas a wearable BCI may prioritize channel configuration coverage and inference throughput. Accordingly, the benchmark does not impose a universal ranking across all dimensions. Instead, it provides normalized model profiles that allow users to select models according to the requirements of a particular task or deployment scenario.

Formally, let $z_{m,j}\in[0,1]$ denote the normalized score of model $m$ on dimension $j$, where a larger value indicates better performance. Given a deployment scenario with non-negative dimension weights $w_j$ satisfying $\sum_j w_j=1$, the overall utility of model $m$ can be defined as

\begin{equation}
U_m(\mathbf{w})
=
\sum_{j=1}^{8} w_j z_{m,j}.
\label{eq:model_profile_utility}
\end{equation}

The weights $\mathbf{w}$ are application-dependent and are not fixed by the benchmark. Users may therefore construct task-specific model rankings by assigning greater weights to the dimensions that matter most for their intended deployment. Hard requirements, such as a minimum channel coverage or a maximum inference latency, can additionally be imposed as feasibility constraints before computing $U_m(\mathbf{w})$.

\begin{figure}[htbp]
    \centering
    \includegraphics[width=\textwidth]{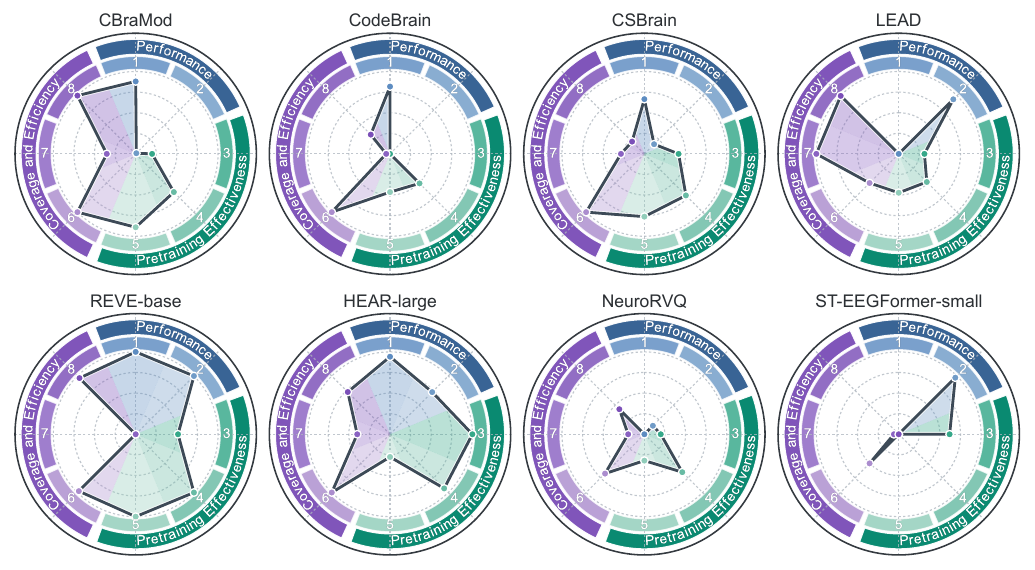}
    \caption{
    Multidimensional evaluation profiles of representative EEG FMs.
    }
    \label{fig:model_profiles}
\end{figure}

Figure~\ref{fig:model_profiles} visualizes the normalized multidimensional profiles of eight representative EEG FMs. The profiles illustrate complementary strengths rather than identify a universally optimal model. For example, \textit{REVE-base} achieves strong average rankings across bipolar and non-bipolar tasks and performs well under limited labeled-data budgets, whereas \textit{HEAR-large} exhibits a substantial final-epoch pretraining gain. \textit{CBraMod} provides broad channel configuration coverage and high inference throughput, while \textit{LEAD} achieves high parameter efficiency despite supporting only non-bipolar configurations. These differences demonstrate why EEG-FM selection should account jointly for task requirements, labeled-data availability, channel configuration, and computational constraints.

\section{Conclusion}

We presented \benchname{}, an open and extensible benchmark for evaluating EEG FMs across heterogeneous tasks, datasets, model architectures, and input configurations. By combining broad comparisons with architecture-matched and fixed-architecture experiments, the benchmark provides a more differentiated view of EEG FM capability than a single performance ranking. The results suggest that the value of pretraining is reflected in both transfer performance and adaptation dynamics, whereas model parameter count alone is insufficient to predict downstream utility. Pretraining data scale and compatibility with the available channel configuration emerge as additional factors that strongly influence practical performance.

These findings also indicate that there is no universally optimal EEG FM. Model selection should depend on the deployment context, including task requirements, labeled-data availability, channel configuration, and computational budget. The multidimensional profiles introduced in \benchname{} provide a basis for making such choices by assigning different importance to performance, pretraining effectiveness, coverage, and efficiency according to the target application.

Several directions remain open. The current model size analysis is observational because existing EEG FMs differ in architecture, pretraining data, optimization procedures, and data quality. More controlled scaling studies are therefore needed to isolate the effects of parameter count and to determine how architecture and pretraining interact. Likewise, our channel configuration experiments focus on structured, task-informed electrode reduction; future work should examine arbitrary channel dropout, changes in referencing schemes, and broader acquisition-device variation. Extending fixed-architecture data scaling experiments to additional model families, datasets, and pretraining objectives would further test the generality of the observed trends.

To support this research agenda, we release the \benchname{} benchmarking framework and online leaderboard. These resources provide standardized interfaces and reporting procedures for integrating new models, datasets, and protocols, enabling reproducible evaluation and cumulative progress toward more capable and practically deployable EEG FMs.

\bibliography{main}

\appendix
\input{appendix_files/appendix_body_setup.tex}

\newpage

\renewcommand*{\theHtable}{app.\arabic{table}}
\renewcommand*{\theHfigure}{app.\arabic{figure}}

\section*{\centering Appendix}
\addcontentsline{toc}{section}{Appendix}
\vspace{3mm}

\section{Content}
\label{app:content}

The appendix is organized as follows:
\begin{enumerate}[leftmargin=*]
  \item \textbf{Datasets and tasks.} Section~\ref{app:datasets} lists the datasets and defines the downstream tasks evaluated in \benchname{}.
  \item \textbf{Model introduction.} Section~\ref{app:models} lists the evaluated EEG FMs and supervised baselines in \benchname{}.
  \item \textbf{Preprocessing pipeline.} Section~\ref{app:preprocessing} presents the preprocessing pipelines of the evaluated EEG FMs and supervised baselines.
  \item \textbf{Supplementary experimental results.} Section~\ref{app:supplementary_experimental_results} presents supplementary results for Protocol I, Protocol II, Protocol III, and Protocol V.
\end{enumerate}

\section{Dataset and Task Coverage in \benchname{}}
\label{app:datasets}

\begingroup
\scriptsize
\setlength{\tabcolsep}{4pt}
\renewcommand{\arraystretch}{0.96}
\begin{longtable}{@{}L{0.19000\linewidth}C{0.07000\linewidth}L{0.25000\linewidth}C{0.05500\linewidth}C{0.06600\linewidth}C{0.06600\linewidth}C{0.08600\linewidth}C{0.07500\linewidth}@{}}
  \caption{Datasets and downstream task definitions in \benchname{}. Task IDs are dataset-specific.}
  \label{app:tab:dataset_tasks}\\
  \toprule
  Dataset & Task ID & Task & Classes & \mbox{tmin (s)} & \mbox{tmax (s)} & \mbox{duration (s)} & Channels \\
  \midrule
  \endfirsthead
  \toprule
  Dataset & Task ID & Task & Classes & \mbox{tmin (s)} & \mbox{tmax (s)} & \mbox{duration (s)} & Channels \\
  \midrule
  \endhead
  \midrule
  \multicolumn{8}{r}{\textit{Continued on next page}} \\
  \endfoot
  \bottomrule
  \endlastfoot
  \multicolumn{8}{@{}l}{\textbf{MI}} \\
  \addlinespace[0.5pt]
  BCIC-IV-1~\cite{blankertz2007bciciv1} & Task 1 & Left/Right Hand MI & 2 & 0.5 & 2.5 & 2.0 & 55 \\
   & Task 2 & Left Hand vs Foot MI & 2 & 0.5 & 2.5 & 2.0 & 55 \\
  \addlinespace[1pt]
  BCIC-IV-2A~\cite{brunner2008bcic_iv_a} & Task 1 & Left Hand/Right Hand/Feet/Tongue MI & 4 & 0.5 & 2.5 & 2.0 & 22 \\
   & Task 2 & Left/Right Hand MI & 2 & 0.5 & 2.5 & 2.0 & 22 \\
   & Task 3 & Hand vs Feet MI & 2 & 0.5 & 2.5 & 2.0 & 22 \\
  \addlinespace[1pt]
  BCIC-IV-2B~\cite{leeb2008bcic_iv_2b} & Task 1 & Left/Right Hand MI & 2 & -0.5 & 4.0 & 4.5 & 3 \\
  \addlinespace[1pt]
  PhysioNet-MI~\cite{schalk2004physionet_mi} & Task 1 & Left Fist/Right Fist/Both Fists/Feet MI & 4 & 0.0 & 4.0 & 4.0 & 64 \\
   & Task 2 & Left/Right Fist MI & 2 & 0.0 & 4.0 & 4.0 & 64 \\
   & Task 3 & Hand vs Feet MI & 2 & 0.0 & 4.0 & 4.0 & 64 \\
  \addlinespace[1pt]
  Large-MI~\cite{kaya2018largemi} & Task 1 & All & 11 & 0.0 & 0.85 & 0.85 & 21--22 \\
   & Task 2 & Left/Right Hand MI & 2 & 0.0 & 3.0 & 3.0 & 21--22 \\
   & Task 3 & Left/Right Hand MI + Passive & 3 & 0.0 & 3.0 & 3.0 & 21--22 \\
   & Task 4 & Left/Right Hand and Left/Right Leg MI & 4 & 0.0 & 3.0 & 3.0 & 21--22 \\
   & Task 5 & Hand vs Leg MI & 2 & 0.0 & 3.0 & 3.0 & 21--22 \\
   & Task 6 & Left/Right Hand, Left/Right Leg, and Tongue MI & 5 & 0.0 & 3.0 & 3.0 & 21--22 \\
   & Task 7 & Individual Fingers & 5 & 0.0 & 3.0 & 3.0 & 21--22 \\
  \addlinespace[1pt]
  SHU-MI~\cite{ma2022shu_mi} & Task 1 & Left/Right Hand MI & 2 & 0.0 & 4.0 & 4.0 & 32 \\
  \addlinespace[1pt]
  Stroke-MI~\cite{liu2024strokemi} & Task 1 & Left/Right Hand MI & 2 & 1.0 & 4.0 & 3.0 & 30 \\
  \addlinespace[1pt]
  Meng2019~\cite{meng2019exploring} & Task 1 & Cursor Control & 2 & 1.5 & 4.5 & 3.0 & 58 \\
   & Task 2 & Cursor Control + Rest & 3 & 1.5 & 4.5 & 3.0 & 58 \\
   & Task 3 & Cursor Movement Detection & 2 & 1.5 & 4.5 & 3.0 & 58 \\
  \addlinespace[1pt]
  HGD~\cite{schirrmeister2017deepconvnet} & Task 1 & Left/Right Hand/Foot & 3 & -0.5 & 4.0 & 4.5 & 128--133 \\
   & Task 2 & Left/Right Hand MI & 2 & -0.5 & 4.0 & 4.5 & 128--133 \\
   & Task 3 & Hand vs Feet MI & 2 & -0.5 & 4.0 & 4.5 & 128--133 \\
  \addlinespace[1pt]
  OpenBMI~\cite{lee2019openbmi} & Task 1 & Left/Right Hand MI & 2 & 0.0 & 4.0 & 4.0 & 62 \\
  \midrule
  \multicolumn{8}{@{}l}{\textbf{Sleep}} \\
  \addlinespace[0.5pt]
  CAP-Sleep~\cite{terzano2001capsleep} & Task 1 & W/S1/S2/S3/S4/REM & 6 & 0.0 & 30.0 & 30.0 & 5 \\
   & Task 2 & Sleep/Wake Detection & 2 & 0.0 & 30.0 & 30.0 & 5 \\
   & Task 3 & Wake/NREM/REM & 3 & 0.0 & 30.0 & 30.0 & 5 \\
   & Task 4 & CAP A-Phase vs B-Phase & 2 & 0.0 & 14.0 & 14.0 & 5 \\
   & Task 5 & W/S1/S2/S3/S4/REM & 6 & 0.0 & 4.0 & 4.0 & 5 \\
   & Task 6 & W/S1/S2-S3/S4/REM & 5 & 0.0 & 30.0 & 30.0 & 5 \\
   & Task 7 & W/S1/S2-S3/S4/REM & 5 & 0.0 & 4.0 & 4.0 & 5 \\
   & Task 8 & Sleep/Wake Detection & 2 & 0.0 & 4.0 & 4.0 & 5 \\
   & Task 9 & Wake/NREM/REM & 3 & 0.0 & 4.0 & 4.0 & 5 \\
   & Task 10 & A1/A2/A3 4s & 3 & 0.0 & 4.0 & 4.0 & 5 \\
  \addlinespace[1pt]
  HMC-Sleep~\cite{alvarez2021hmcsleep} & Task 1 & Sleep/Wake Detection & 2 & 0.0 & 30.0 & 30.0 & 4--8 \\
   & Task 2 & Wake/NREM/REM & 3 & 0.0 & 30.0 & 30.0 & 4--8 \\
   & Task 3 & W/N1/N2/N3/REM & 5 & 0.0 & 30.0 & 30.0 & 4--8 \\
  \addlinespace[1pt]
  UCDDB-Sleep~\cite{goldberger2000physionet} & Task 1 & W/S1/S2/S3/S4/REM & 6 & 0.0 & 30.0 & 30.0 & 2 \\
   & Task 2 & W/S1/S2/S3-S4/REM & 5 & 0.0 & 30.0 & 30.0 & 2 \\
   & Task 3 & Sleep/Wake Detection & 2 & 0.0 & 30.0 & 30.0 & 2 \\
   & Task 4 & Wake/NREM/REM & 3 & 0.0 & 30.0 & 30.0 & 2 \\
  \addlinespace[1pt]
  Sleep-EDF~\cite{kemp2000analysis} & Task 1 & W/S1/S2-S3/S4/REM & 5 & 0.0 & 30.0 & 30.0 & 2 \\
   & Task 2 & Sleep/Wake Detection & 2 & 0.0 & 30.0 & 30.0 & 2 \\
   & Task 3 & Wake/NREM/REM & 3 & 0.0 & 30.0 & 30.0 & 2 \\
  \midrule
  \multicolumn{8}{@{}l}{\textbf{Clinical}} \\
  \addlinespace[0.5pt]
  TUSL~\cite{obeid2016tuh_eeg} & Task 1 & Background Slowing Detection & 2 & 0.0 & 5.0 & 5.0 & 21 \\
   & Task 2 & Background/Slowing/Seizure Classification & 3 & 0.0 & 5.0 & 5.0 & 21 \\
  \addlinespace[1pt]
  TUEV~\cite{obeid2016tuh_eeg} & Task 1 & Background/Spike and Sharp Wave/GPED/PLED & 4 & -0.5 & 1.0 & 1.5 & 21 \\
   & Task 2 & Background/Epileptiform Event & 2 & -0.5 & 1.0 & 1.5 & 21 \\
  \addlinespace[1pt]
  MigraineDB~\cite{chamanzar2020migrainedb} & Task 1 & Resting-state Migraine Detection & 2 & 0.0 & 5.0 & 5.0 & 130--144 \\
   & Task 2 & SSVEP Migraine Detection & 2 & 0.0 & 2.0 & 2.0 & 130--144 \\
   & Task 3 & SSAEP Migraine Detection & 2 & 0.0 & 2.0 & 2.0 & 130--144 \\
   & Task 4 & SSVEP Frequency Discrimination & 2 & 0.0 & 2.0 & 2.0 & 130--144 \\
   & Task 5 & SSAEP Frequency Discrimination & 2 & 0.0 & 2.0 & 2.0 & 130--144 \\
   & Task 6 & 4-Hz SSVEP Migraine Detection & 2 & 0.0 & 2.0 & 2.0 & 130--144 \\
   & Task 7 & 6-Hz SSVEP Migraine Detection & 2 & 0.0 & 2.0 & 2.0 & 130--144 \\
   & Task 8 & 4-Hz SSAEP Migraine Detection & 2 & 0.0 & 2.0 & 2.0 & 130--144 \\
   & Task 9 & 6-Hz SSAEP Migraine Detection & 2 & 0.0 & 2.0 & 2.0 & 130--144 \\
  \addlinespace[1pt]
  TUAB~\cite{obeid2016tuh_eeg} & Task 1 & Abnormal EEG Detection & 2 & 0.0 & 5.0 & 5.0 & --- \\
  \addlinespace[1pt]
  TUEP~\cite{veloso2017tuep} & Task 1 & Epilepsy Detection & 2 & 0.0 & 5.0 & 5.0 & --- \\
  \addlinespace[1pt]
  TUSZ~\cite{obeid2016tuh_eeg} & Task 1 & Seizure Detection & 2 & 0.0 & 2.0 & 2.0 & --- \\
   & Task 2 & Seizure Type & 9 & 0.0 & 2.0 & 2.0 & --- \\
  \addlinespace[1pt]
  TUAR~\cite{hamid2020tuar} & Task 1 & Artifact Detection & 2 & 0.0 & 1.0 & 1.0 & --- \\
   & Task 2 & Artifact Type & 14 & 0.0 & 1.0 & 1.0 & --- \\
  \addlinespace[1pt]
  CHB-MIT~\cite{shoeb2009chbmit} & Task 1 & Seizure Detection & 2 & 0.0 & 2.0 & 2.0 & --- \\
  \midrule
  \multicolumn{8}{@{}l}{\textbf{Cognitive}} \\
  \addlinespace[0.5pt]
  EEGMat~\cite{zyma2019eegmat} & Task 1 & Mental Arithmetic vs Rest & 2 & 0.0 & 4.0 & 4.0 & 19 \\
  \midrule
  \multicolumn{8}{@{}l}{\textbf{Auditory}} \\
  \addlinespace[0.5pt]
  KUL-AAD~\cite{biesmans2016kulaad} & Task 1 & Attended Side Decoding & 2 & 0.0 & 5.0 & 5.0 & 64 \\
  \addlinespace[1pt]
  DTU-AAD~\cite{fuglsang2017dtuaad} & Task 1 & Left/Right Attention Decoding & 2 & 0.0 & 5.0 & 5.0 & 64 \\
   & Task 2 & Gender Attention Decoding & 2 & 0.0 & 5.0 & 5.0 & 64 \\
   & Task 3 & Gender $\times$ Left/Right Attention Decoding & 4 & 0.0 & 5.0 & 5.0 & 64 \\
  \midrule
  \multicolumn{8}{@{}l}{\textbf{ERP}} \\
  \addlinespace[0.5pt]
  ERP-BCI~\cite{citi2010physionetp300} & Task 1 & P300 Target Detection & 2 & -0.1 & 0.8 & 0.9 & 64 \\
  \addlinespace[1pt]
  Kaggle-INRIA~\cite{margaux2012kaggleERN} & Task 1 & P300 Feedback Correctness & 2 & 0.0 & 1.0 & 1.0 & 56 \\
  \midrule
  \multicolumn{8}{@{}l}{\textbf{Emotion}} \\
  \addlinespace[0.5pt]
  SEED~\cite{zheng2015seed} & Task 1 & Negative/Neutral/Positive & 3 & 0.0 & 4.0 & 4.0 & 62 \\
  \addlinespace[1pt]
  FACED~\cite{chen2023faced} & Task 1 & Emotion Recognition & 9 & 0.0 & 10.0 & 10.0 & 30 \\
\end{longtable}
\endgroup

\section{Models evaluated in \benchname{}}
\label{app:models}

In this section, the EEG FM models and supervised baselines evaluated in \benchname{} are listed.

\begingroup
\setlength{\tabcolsep}{3pt}
\renewcommand{\arraystretch}{0.94}
\begin{table}[H]
  \caption{Evaluated EEG FM models.}
  \label{tab:fms}
  \centering
  \scriptsize
  \setlength{\tabcolsep}{6pt}
  \begin{tabular*}{\linewidth}{@{\extracolsep{\fill}}llllccl@{}}
    \toprule
    Model & Year & Variant & Parameters & Input & Sampling rate (Hz) & Channel input strategy \\
    \midrule
    BENDR~\cite{kostas2021bendr} & 2021 & --- & 157.0M & raw & 256 & Channel-constrained \\
    \addlinespace[2pt]
    \multirow{3}{*}{BIOT~\cite{yang2023biot}} & \multirow{3}{*}{2023} & shhs-prest-18ch & 3.3M & \multirow{3}{*}{raw} & \multirow{3}{*}{200} & \multirow{3}{*}{Channel-constrained} \\
     &  & prest-16ch & 3.3M &  &  &  \\
     &  & six-datasets-18ch & 3.3M &  &  &  \\
    \addlinespace[2pt]
    EEGPT~\cite{wang2024eegpt} & 2024 & --- & 25.3M & patched & 256 & Channel-constrained \\
    \addlinespace[2pt]
    LaBraM~\cite{jiang2024large} & 2024 & base & 5.8M & patched & 200 & Channel-constrained \\
    \addlinespace[2pt]
    NeuroGPT~\cite{cui2024neurogpt} & 2024 & --- & 79.5M & patched & 250 & Channel-constrained \\
    \addlinespace[2pt]
    CBraMod~\cite{wang2025cbramod} & 2025 & --- & 4.0M & patched & 200 & Channel-constrained \\
    \addlinespace[2pt]
	    CSBrain~\cite{zhou2025csbrain} & 2025 & --- & 8.90M & patched & 200 & Channel-constrained \\
	    \addlinespace[2pt]
	    EEGMamba~\cite{wang2025eegmamba} & 2025 & --- & 3.3M & patched & 200 & Channel-constrained \\
	    \addlinespace[2pt]
	    \multirow{2}{*}{HEAR~\cite{chen2025hear}} & \multirow{2}{*}{2025} & base & 3.1M & \multirow{2}{*}{patched} & \multirow{2}{*}{200} & \multirow{2}{*}{Channel-flexible} \\
	     &  & large & 6.1M &  &  &  \\
	    \addlinespace[2pt]
	    LEAD~\cite{wang2025lead} & 2025 & base & 3.41M & raw & 128 & Channel-flexible \\
    \addlinespace[2pt]
    \multirow{3}{*}{LUNA~\cite{doner2025luna}} & \multirow{3}{*}{2025} & base & 7.0M & \multirow{3}{*}{raw} & \multirow{3}{*}{256} & \multirow{3}{*}{Channel-flexible} \\
     &  & large & 43.0M &  &  &  \\
     &  & huge & 311.0M &  &  &  \\
    \addlinespace[2pt]
    mdJPT~\cite{zhang2025multi} & 2025 & --- & 1.0M & patched & 125 & Channel-constrained \\
    \addlinespace[2pt]
    NeuroRVQ~\cite{barmpas2025neurorvq} & 2025 & --- & 5.9M & patched & 200 & Channel-constrained \\
    \addlinespace[2pt]
    \multirow{2}{*}{REVE~\cite{el2025reve}} & \multirow{2}{*}{2025} & base & 69.0M & \multirow{2}{*}{raw} & \multirow{2}{*}{200} & \multirow{2}{*}{Channel-flexible} \\
     &  & large & 408.0M &  &  &  \\
    \addlinespace[2pt]
    CodeBrain~\cite{ma2025codebrain} & 2026 & --- & 15.2M & patched & 200 & Channel-constrained \\
    \addlinespace[2pt]
    \multirow{3}{*}{ST-EEGFormer~\cite{yang2026areeeg}} & \multirow{3}{*}{2026} & small & 32.7M & \multirow{3}{*}{raw} & \multirow{3}{*}{128} & \multirow{3}{*}{Channel-constrained} \\
     &  & base & 110.9M &  &  &  \\
     &  & large & 328.4M &  &  &  \\
    \addlinespace[2pt]
    \multirow{3}{*}{SleepFM~\cite{thapa2026sleepfm}} & \multirow{3}{*}{2026} & sleeping-staging & 6.12M & \multirow{3}{*}{raw} & \multirow{3}{*}{128} & \multirow{3}{*}{Channel-flexible} \\
     &  & diagnosis & 5.80M &  &  &  \\
     &  & upstream & 4.83M &  &  &  \\
    \addlinespace[2pt]
    SleepGPT~\cite{huang2026sleepgpt} & 2026 & large & 134M & raw & 100 & Channel-constrained \\
    \addlinespace[2pt]
    TFM~\cite{pradeepkumar2026tfm} & 2026 & CHBMIT & 1.89M & raw & 200 & Channel-constrained \\
    \addlinespace[2pt]
    Uni-DMFM~\cite{chen2026unidmfm} & 2026 & base & 29.42M & patched & 200 & Channel-flexible \\
    \bottomrule
  \end{tabular*}
\end{table}
\endgroup

\begin{table}[H]
  \caption{Supervised baseline models evaluated in \benchname{}.}
  \label{tab:baselines}
  \centering
  \scriptsize
  \begin{minipage}[t]{0.47\linewidth}
  \centering
  \resizebox{\linewidth}{!}{%
  \begin{tabular}[t]{lllc}
    \toprule
    Model & Year & Architecture & Domain \\
    \midrule
    EEGNet~\cite{lawhern2018eegnet} & 2018 & CNN & General \\
    ShallowConvNet~\cite{schirrmeister2017deepconvnet} & 2017 & CNN & General \\
    DeepConvNet~\cite{schirrmeister2017deepconvnet} & 2017 & CNN & General \\
    EEGConformer~\cite{song2022eegconformer} & 2023 & CNN+Transformer & General \\
    ATCNet~\cite{altaheri2022atcnet} & 2023 & CNN+Attention+TCN & MI \\
    CTNet~\cite{zhao2024ctnet} & 2024 & CNN+Transformer & MI \\
    MSVTNet~\cite{liu2024msvtnet} & 2024 & CNN+ViT & MI \\
    EEGInceptionERP~\cite{santamaria2020eeginceptionerp} & 2020 & Inception & ERP \\
    DBConformer~\cite{wang2025dbconformer} & 2025 & CNN+Transformer & General \\
    FBCNet~\cite{mane2021fbcnet} & 2021 & CNN & MI \\
    IFNet~\cite{wang2023ifnet} & 2023 & CNN & MI \\
    EEGWaveNet~\cite{thuwajit2021eegwavenet} & 2022 & CNN & Epilepsy \\
    SlimSeiz~\cite{lu2025slimseiz} & 2025 & Mamba & Epilepsy \\
    \bottomrule
  \end{tabular}%
  }
  \end{minipage}%
  \hspace{0.06\linewidth}%
  \begin{minipage}[t]{0.47\linewidth}
  \centering
  \resizebox{\linewidth}{!}{%
  \begin{tabular}[t]{lllc}
    \toprule
    Model & Year & Architecture & Domain \\
    \midrule
    MSCFormer~\cite{zhao2025mscformer} & 2025 & CNN+Transformer & MI \\
    TMSA-Net~\cite{zhao2025tmsanet} & 2025 & CNN & MI \\
    ADFCNN~\cite{tao2023adfcnn} & 2024 & CNN & MI \\
    EEGNeX~\cite{chen2024eegnex} & 2024 & CNN & General \\
    BiDANN~\cite{li2018bidann} & 2021 & Adversarial Network & Emotion \\
    TSception~\cite{ding2022tsception} & 2023 & Inception & Emotion \\
    HSLT~\cite{wang2022hslt} & 2022 & Transformer & Emotion \\
    DeepSleepNet~\cite{supratak2017deepsleepnet} & 2017 & CNN+BiLSTM & Sleep \\
    SleepEEGNet~\cite{mousavi2019sleepeegnet} & 2019 & CNN+BiLSTM & Sleep \\
    SleepTransformer~\cite{phan2022sleeptransformer} & 2022 & Transformer & Sleep \\
    U-Sleep~\cite{perslev2021usleep} & 2021 & U-Net & Sleep \\
    YASA~\cite{vallat2021yasa} & 2021 & Spectral & Sleep \\
    \bottomrule
  \end{tabular}%
  }
  \end{minipage}
\end{table}

\FloatBarrier

\section{Preprocessing Pipelines of EEG FMs and Supervised Baselines}
\label{app:preprocessing}
Table~\ref{tab:native_protocols} presents the preprocessing parameters of each EEG FM and the unified classical preprocessing pipeline used by supervised baselines.

\begin{table}[H]
  \caption{Preprocessing pipelines of the evaluated EEG FMs and the supervised baselines.}
  \label{tab:native_protocols}
  \centering
  \scriptsize
  \textbf{(a) EEG FM preprocessing pipelines}\par\vspace{2pt}
  \resizebox{\linewidth}{!}{%
  \begin{tabular}{lllllll}
    \toprule
    Model & Sampling rate (Hz) & Band-pass (Hz) & Notch (Hz) & Normalization & Overlap / Stride & Input unit \\
    \midrule
    BENDR & 256 & --- & --- & Scale to $[-1,1]$ & No overlap & Raw \\
    BIOT & 200 & --- & --- & 95th percentile & Continuous & Patch (1s) \\
    CBraMod & 200 & 0.3--75 & 60 & $\div100\,\mu$V, reject >100$\mu$V & No overlap & Patch (1s / 200) \\
    CodeBrain & 200 & 0.3--75 & 60 & $\div100$, reject >100$\mu$V & No overlap & Patch (1s / 200) \\
    CSBrain & 200 & 0.3--75 & 60 & $\div100\,\mu$V & No overlap & Patch (1s / 200, multiscale) \\
    EEGMamba & 200 & 0.3--75 & 50/60 & $\div100\,\mu$V & No overlap & Patch (1s / 200) \\
    EEGPT & 256 & Task-dependent$^1$ & --- & Unified to mV & --- & Patch (0.25s / 64) \\
    HEAR & 200 & 0.1--75 & 50 & $\div0.1\,$mV & Continuous & Patch (1s / 200) \\
    LaBraM & 200 & 0.1--75 & 50 & $\div0.1\,$mV & Continuous & Patch (1s / 200) \\
    LEAD & 128 & 0.5--45 & --- & Z-score & 50\% overlap & Internal (128) \\
    LUNA & 256 & 0.1--75 & 50/60 & Z-score & Continuous & Internal (40) \\
    mdJPT & 125 & 0.5--47 & --- & --- & 2s stride & Internal (patch) \\
    NeuroGPT & 250 & 0.5--100 & 60 & Z-transform & 10\% overlap & Raw \\
    NeuroRVQ & 200 & --- & --- & --- & --- & Patch (multiscale) \\
    REVE & 200 & 0.5--99.5 & --- & Session-wise z-score, $\pm15\sigma$ clip & --- & Patch (1s) \\
    SleepFM & 128 & 64 Hz low-pass & --- & Channel-wise z-score & No overlap & Token (5s / 640) \\
    SleepGPT & 100 & Task-dependent & --- & Z-score & No overlap & Patch (2s / 200) \\
    ST-EEGFormer & 128 & 0.1--64 & --- & Z-score & 0.5s overlap & Internal (16) \\
    TFM (CHBMIT) & 200 & --- & --- & 95th percentile & 0.5s overlap & Patch (1s / 200) \\
    Uni-DMFM & 200 & Task-dependent & Task-dependent & Z-score & No overlap & Patch (1s / 200) \\
    \bottomrule
  \end{tabular}}
  \par\vspace{5pt}
  \textbf{(b) Unified preprocessing pipeline for supervised baselines}\par\vspace{2pt}
  \resizebox{\linewidth}{!}{%
  \begin{tabular}{lllllll}
    \toprule
    Model group & Sampling rate (Hz) & Band-pass (Hz) & Notch (Hz) & Normalization & Overlap / Stride & Input unit \\
    \midrule
    Supervised baselines & 200 & 0.1--75 & 50 & Channel-wise z-score & No overlap & Raw \\
    \bottomrule
  \end{tabular}}
\end{table}

\section{Supplementary Experimental Results}
\label{app:supplementary_experimental_results}

In this section, we present supplementary result tables and figures organized according to the protocol structure used in the main manuscript. The supplementary materials are grouped as follows:
\begin{itemize}[leftmargin=*]
  \item \textbf{Protocol I: Comparative Evaluation of EEG FMs and Supervised Baselines.} Section~\ref{app:overall_benchmark_results} presents the benchmark performance of all the evaluated models across tasks.
  \item \textbf{Protocol II: Pretraining Benefits Across Adaptation Stages and Data Budgets.} Sections~\ref{app:early_transfer_dynamics} and~\ref{app:data_budget_adaptation_results} present paired pretraining-benefit results and fine-tuning data-budget results for Full-FT, Linear Probe, and Scratch-ST settings.
  \item \textbf{Protocol III: The Scaling of Model Sizes.} Section~\ref{app:model_scaling_results} presents Linear Probe and Full-FT results for all evaluated EEG FMs with multiple variants.
  \item \textbf{Protocol V: Robustness to Channel Configuration.} Section~\ref{app:input_space_retention_results} presents channel-configuration definitions and channel robustness results.
\end{itemize}

\subsection{Protocol I: Comparative Evaluation of EEG FMs and Supervised Baselines}
\label{app:overall_benchmark_results}

In this subsection, Tables~\ref {tab:dense_matrix_table_i} to~\ref {tab:dense_matrix_table_i_19} report the result matrices for all models and tasks.

\begingroup
\colorlet{AppBaselineCell}{findingredBg}
\colorlet{AppFMCell}{rqblueBg}

\input{appendix_files/tex/app_tab02_overall_performance_results.tex}
\endgroup

\FloatBarrier

\subsection{Protocol II: Pretraining Benefits Across Adaptation Stages and Data Budgets}
\label{sec:results_pretraining}
\label{app:early_transfer_dynamics}

\begingroup
\colorlet{AppBaselineCell}{findingredBg}
\colorlet{AppFMCell}{rqblueBg}

\subsubsection{Learning Curves for All EEG FMs and Supervised Baselines}
\label{app:learning_curves_all_models}

Figure~\ref{fig:app-warmstart-overall-curves-01} to Figure~\ref{fig:app-warmstart-task-family-model-curves-07} present the training curves of all EEG FMs and baselines by task.

\begin{figure}[H]
  \centering
  \includegraphics[width=\linewidth]{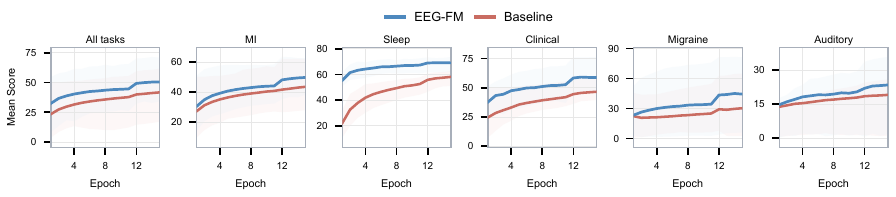}
  \caption{Learning curves for all EEG FMs across all tasks.}
  \label{fig:app-warmstart-overall-curves-01}
\end{figure}

\begin{figure}[H]
  \centering
  \includegraphics[width=\linewidth]{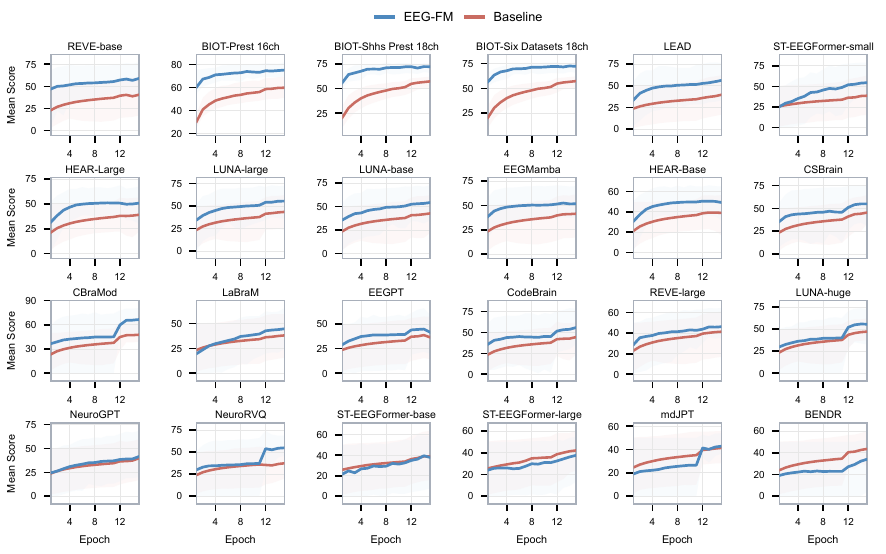}
  \caption{Learning curves for all EEG FMs across all tasks.}
  \label{fig:app-warmstart-representative-model-curves-01}
\end{figure}

\begin{figure}[H]
  \centering
  \includegraphics[width=\linewidth]{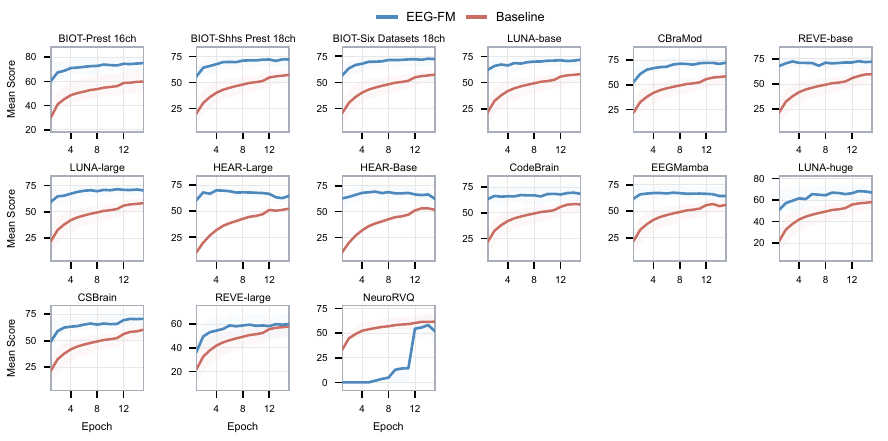}
  \caption{Learning curves for all EEG FMs in Sleep task.}
  \label{fig:app-warmstart-task-family-model-curves-02}
\end{figure}

\begin{figure}[H]
  \centering
  \includegraphics[width=\linewidth]{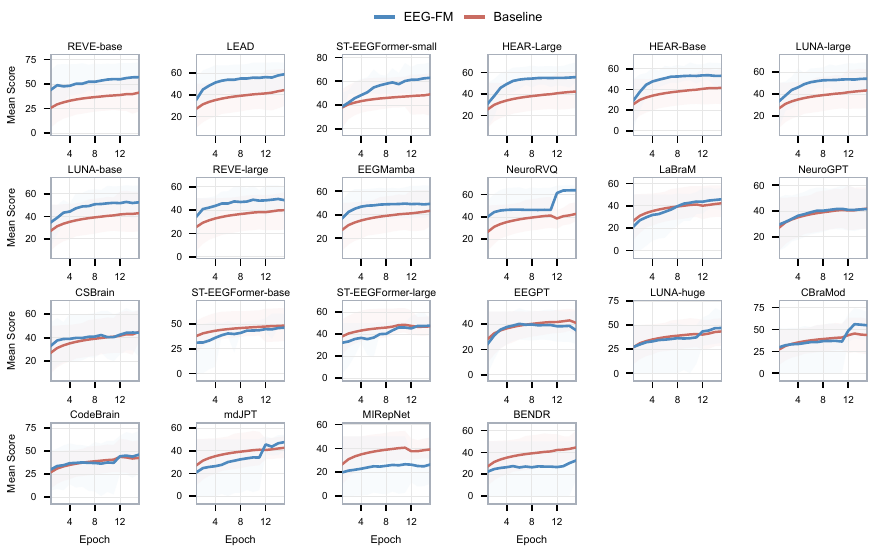}
  \caption{Learning curves for all EEG FMs in MI task.}
  \label{fig:app-warmstart-task-family-model-curves-01}
\end{figure}

\begin{figure}[H]
  \centering
  \includegraphics[width=\linewidth]{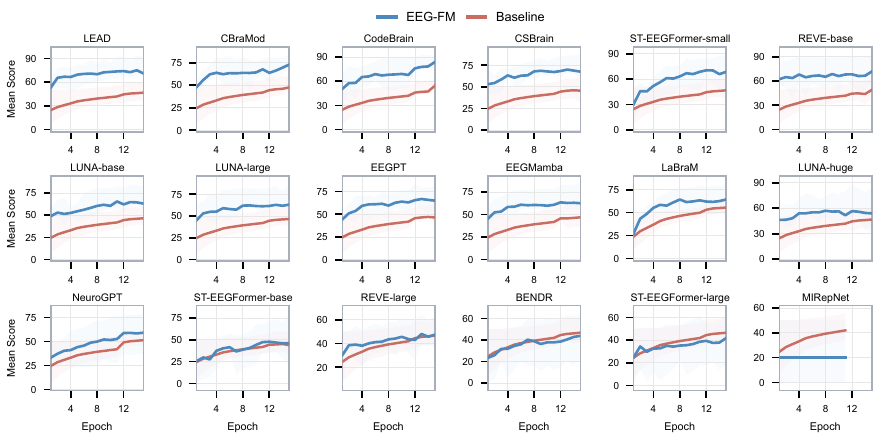}
  \caption{Learning curves for all EEG FMs in Clinical task.}
  \label{fig:app-warmstart-task-family-model-curves-03}
\end{figure}

\begin{figure}[H]
  \centering
  \includegraphics[width=\linewidth]{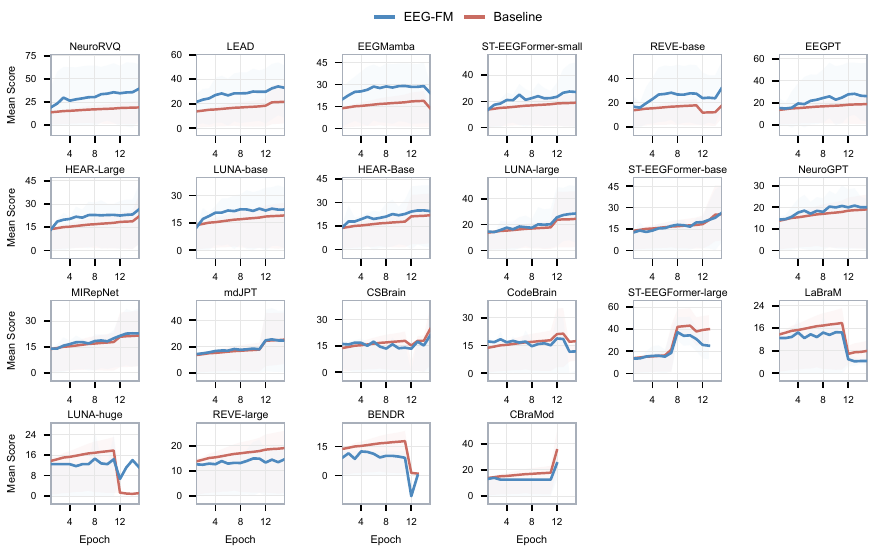}
  \caption{Learning curves for all EEG FMs in Auditory task.}
  \label{fig:app-warmstart-task-family-model-curves-05}
\end{figure}

\begin{figure}[H]
  \centering
  \includegraphics[width=\linewidth]{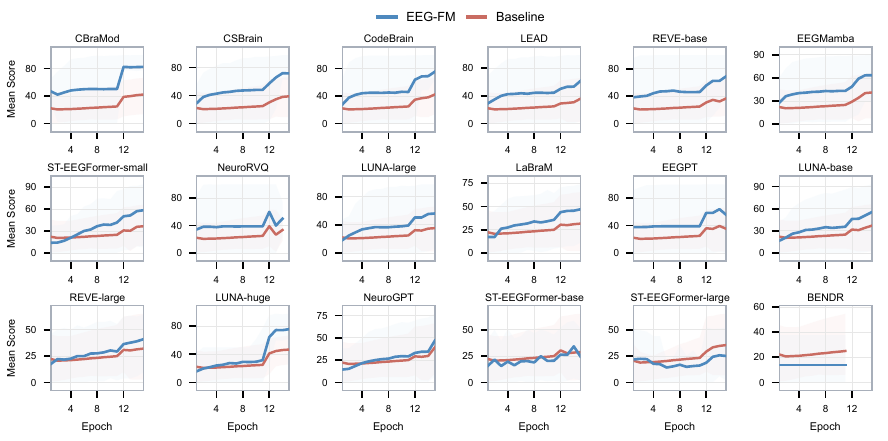}
  \caption{Learning curves for all EEG FMs in Migraine task.}
  \label{fig:app-warmstart-task-family-model-curves-04}
\end{figure}

\begin{figure}[H]
  \centering
  \includegraphics[width=\linewidth]{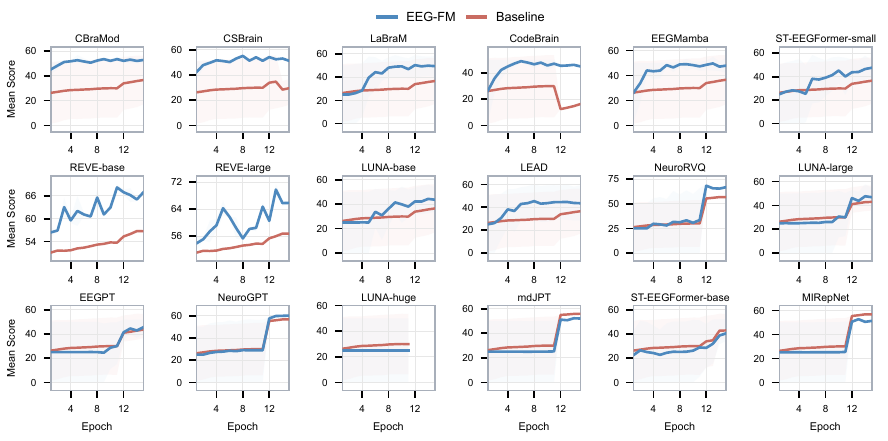}
  \caption{Learning curves for all EEG FMs in ERP task.}
  \label{fig:app-warmstart-task-family-model-curves-06}
\end{figure}

\begin{figure}[H]
  \centering
  \includegraphics[width=\linewidth]{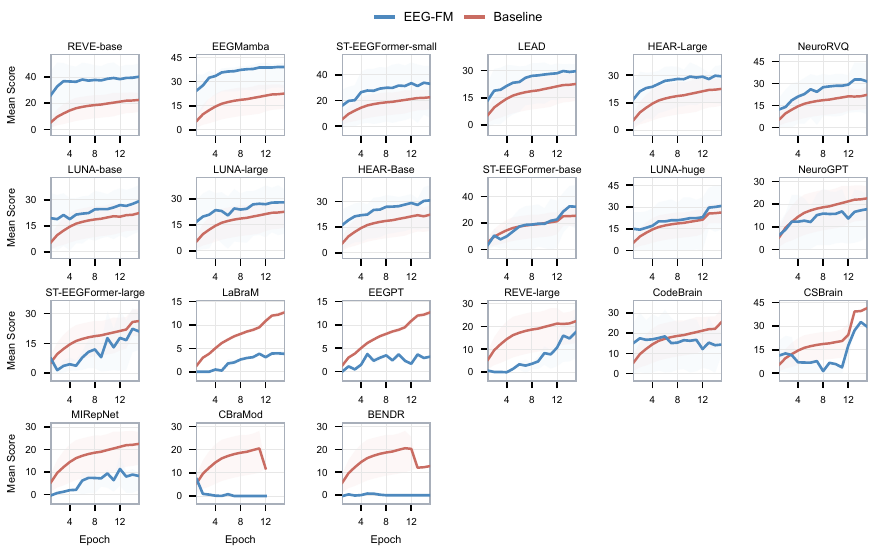}
  \caption{Learning curves for all EEG FMs in Emotion task.}
  \label{fig:app-warmstart-task-family-model-curves-07}
\end{figure}

\FloatBarrier




\subsubsection{Average Learning Curves for EEG FMs and Supervised Baselines}
\label{app:average_learning_curves}

Figures~\ref{fig:app-warmstart-curves-03}--\ref{fig:app-warmstart-curves-09} present the average training curves of all the EEG FMs and baselines by task.

\begingroup
\setlength{\intextsep}{0pt}
\setlength{\abovecaptionskip}{2pt}
\begin{figure}[H]
  \centering
  \includegraphics[width=\linewidth]{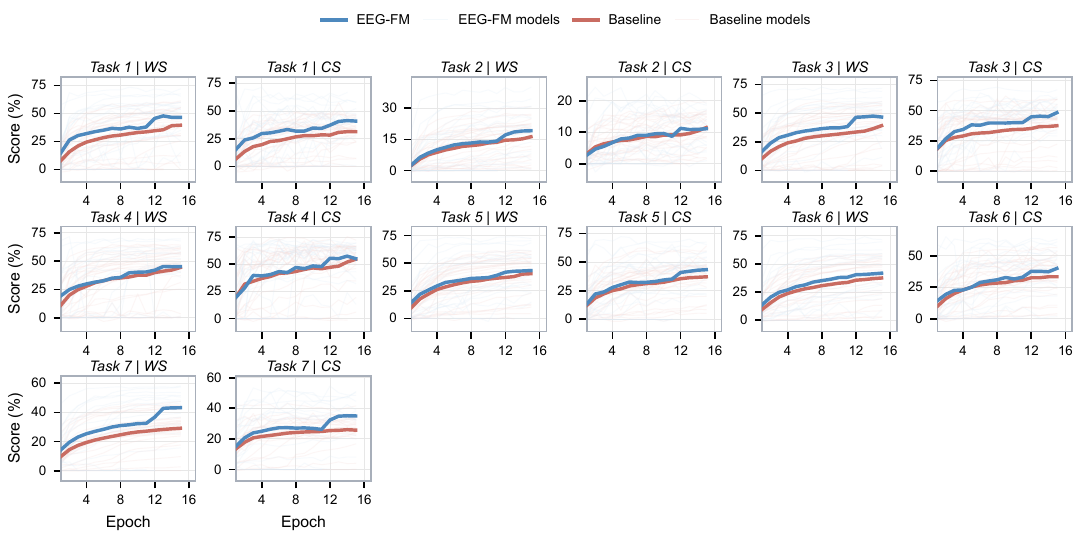}
  \caption{Average learning curves in Large-MI task.}
  \label{fig:app-warmstart-curves-03}
\end{figure}

\begin{figure}[H]
  \centering
  \includegraphics[width=\linewidth]{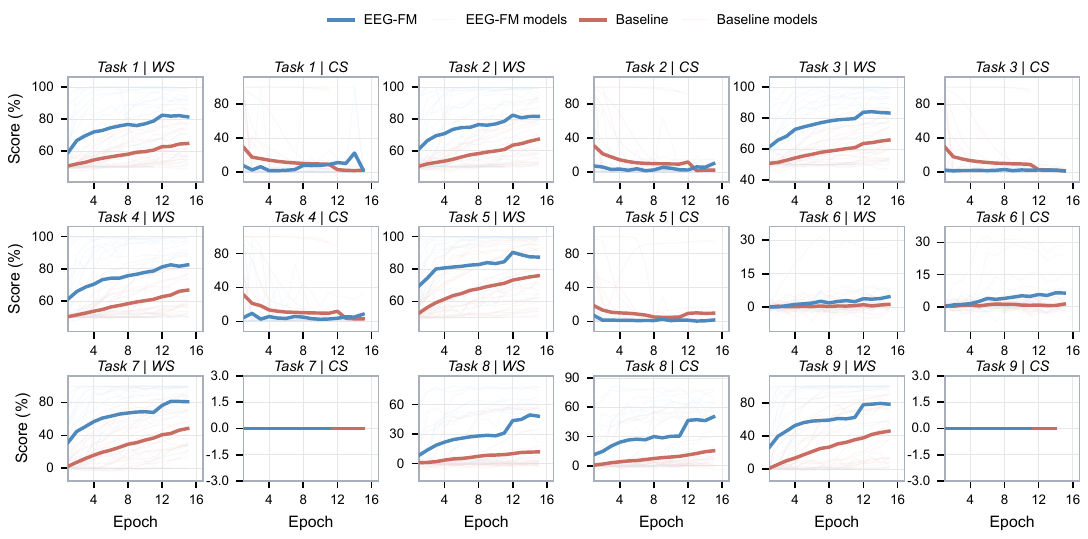}
  \caption{Average learning curves in MigraineDB task.}
  \label{fig:app-warmstart-curves-07}
\end{figure}

\begin{figure}[H]
  \centering
  \includegraphics[width=\linewidth]{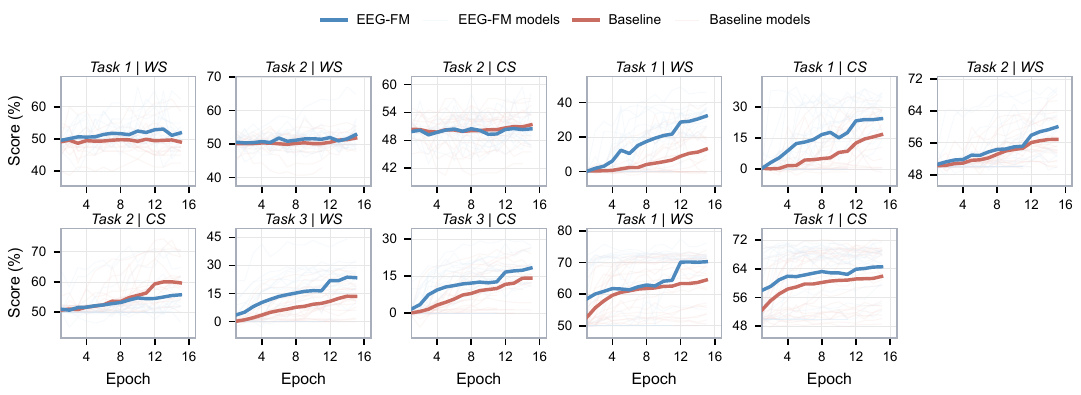}
  \caption{Average learning curves in BCIC-IV-1, BCIC-IV-2A, and BCIC-IV-2B tasks.}
  \label{fig:app-warmstart-curves-01}
\end{figure}

\begin{figure}[H]
  \centering
  \includegraphics[width=\linewidth]{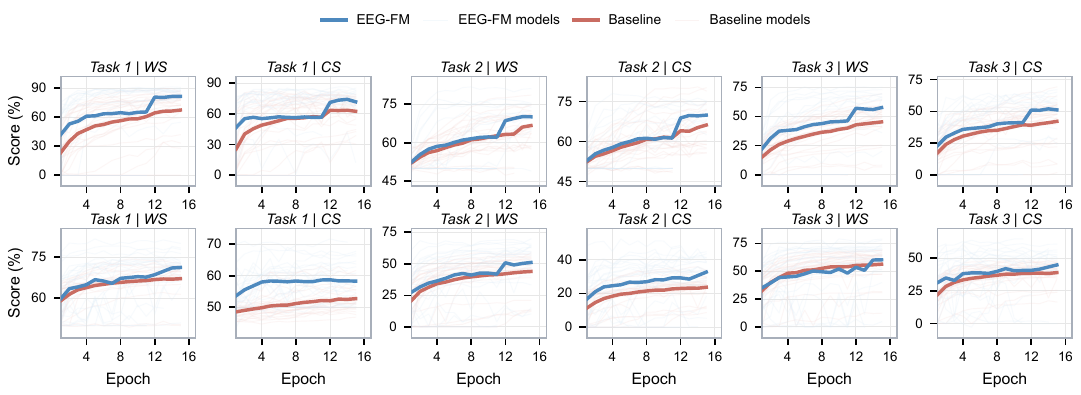}
  \caption{Average learning curves in HGD and Meng2019 tasks.}
  \label{fig:app-warmstart-curves-02}
\end{figure}

\begin{figure}[H]
  \centering
  \includegraphics[width=\linewidth]{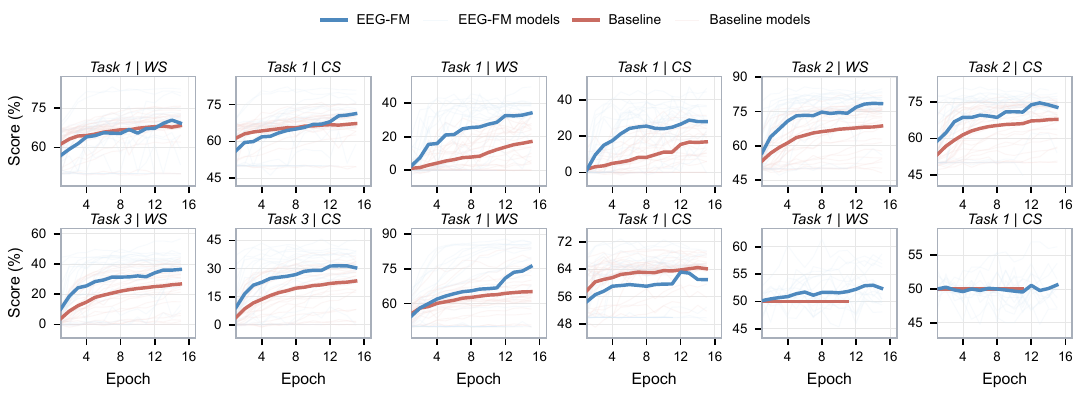}
  \caption{Average learning curves in OpenBMI, PhysioNet-MI, SHU-MI, and Stroke-MI tasks.}
  \label{fig:app-warmstart-curves-04}
\end{figure}

\begin{figure}[H]
  \centering
  \includegraphics[width=\linewidth]{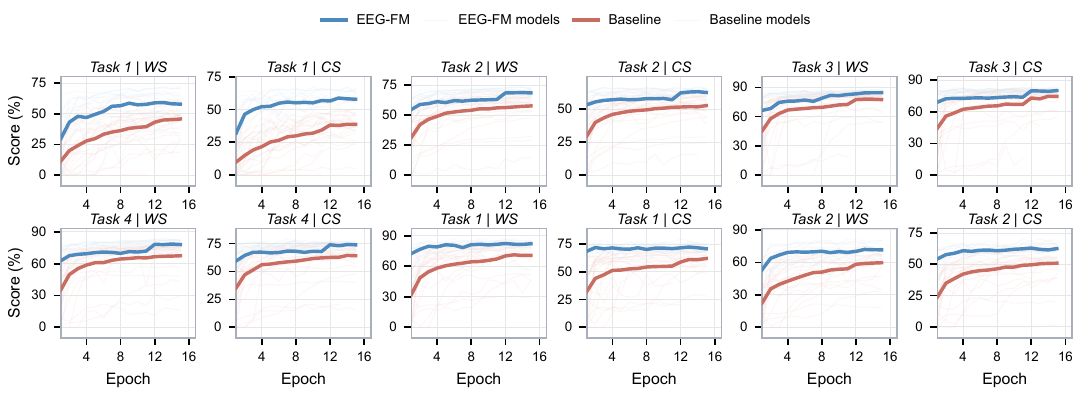}
  \caption{Average learning curves in CAP-Sleep and HMC-Sleep tasks.}
  \label{fig:app-warmstart-curves-05}
\end{figure}

\begin{figure}[H]
  \centering
  \includegraphics[width=\linewidth]{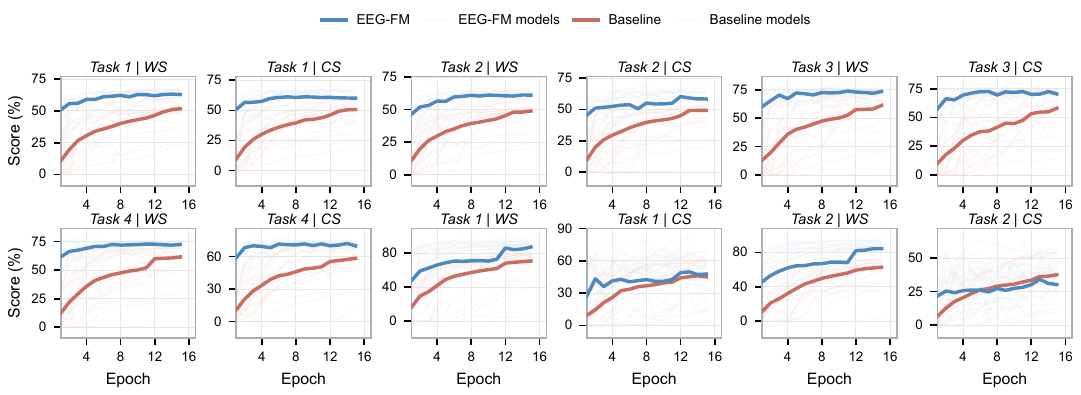}
  \caption{Average learning curves in UCDDB-Sleep and TUEV tasks.}
  \label{fig:app-warmstart-curves-06}
\end{figure}

\begin{figure}[H]
  \centering
  \includegraphics[width=\linewidth]{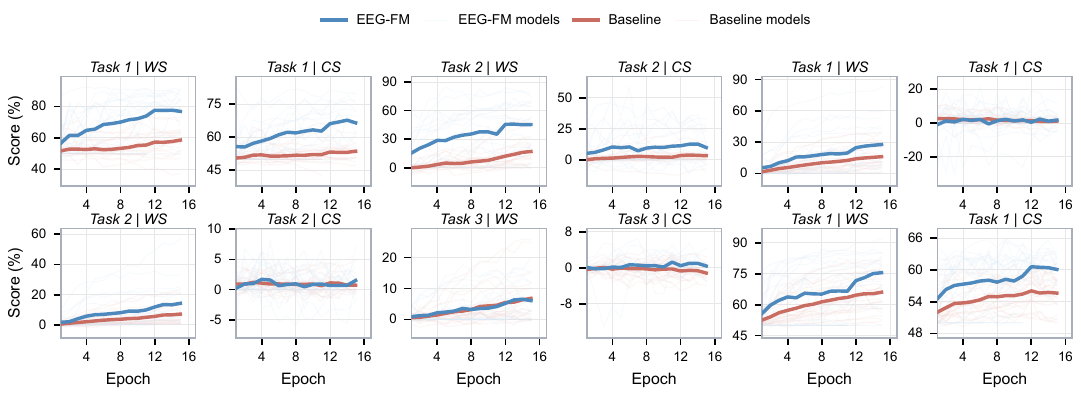}
  \caption{Average learning curves in TUSL, DTU-AAD, and KUL-AAD tasks.}
  \label{fig:app-warmstart-curves-08}
\end{figure}

\begin{figure}[H]
  \centering
  \includegraphics[width=\linewidth]{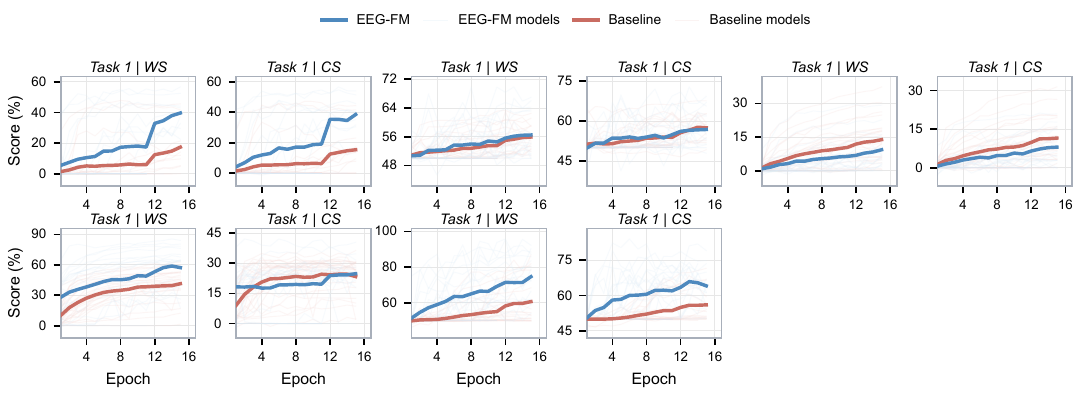}
  \caption{Average learning curves in ERP-BCI, Kaggle-INRIA, FACED, SEED, and EEGMat tasks.}
  \label{fig:app-warmstart-curves-09}
\end{figure}

\FloatBarrier
\endgroup

\subsubsection{Data Budget and Adaptation Strategy Results}
\label{app:data_budget_adaptation_results}

In this section, we present the data budget and adaptation results for EEG FMs and supervised baselines.

\begingroup
\setlength{\LTpre}{8pt}
\setlength{\LTpost}{8pt}
\fontsize{5.2}{5.95}\selectfont
\setlength{\tabcolsep}{0.05pt}
\renewcommand{\arraystretch}{0.94}
\begin{longtable}{@{}L{0.1040\linewidth}C{0.0750\linewidth}C{0.0431\linewidth}C{0.0431\linewidth}C{0.0431\linewidth}C{0.0431\linewidth}C{0.0431\linewidth}C{0.0431\linewidth}C{0.0431\linewidth}C{0.0431\linewidth}C{0.0431\linewidth}C{0.0431\linewidth}C{0.0431\linewidth}C{0.0431\linewidth}C{0.0431\linewidth}C{0.0431\linewidth}C{0.0431\linewidth}C{0.0431\linewidth}C{0.0431\linewidth}C{0.0431\linewidth}@{}}
\caption{Model--task performance by adaptation strategy and fine-tuning data budget for EEGMat and TUSL.}\label{tab:dense_matrix_table_iv}\\
\toprule
\multirow{2}{=}{\raggedright\textbf{Model}} & \multirow{2}{=}{\centering\textbf{Adaptation Strategies}} & \multicolumn{3}{c}{\textbf{2\%}} & \multicolumn{3}{c}{\textbf{5\%}} & \multicolumn{3}{c}{\textbf{10\%}} & \multicolumn{3}{c}{\textbf{25\%}} & \multicolumn{3}{c}{\textbf{50\%}} & \multicolumn{3}{c}{\textbf{100\%}}\\
\cmidrule(lr){3-5}\cmidrule(lr){6-8}\cmidrule(lr){9-11}\cmidrule(lr){12-14}\cmidrule(lr){15-17}\cmidrule(lr){18-20}
 & & B-ACC & Kappa & F1 & B-ACC & Kappa & F1 & B-ACC & Kappa & F1 & B-ACC & Kappa & F1 & B-ACC & Kappa & F1 & B-ACC & Kappa & F1\\
\midrule
\endfirsthead
\toprule
\multirow{2}{=}{\raggedright\textbf{Model}} & \multirow{2}{=}{\centering\textbf{Adaptation Strategies}} & \multicolumn{3}{c}{\textbf{2\%}} & \multicolumn{3}{c}{\textbf{5\%}} & \multicolumn{3}{c}{\textbf{10\%}} & \multicolumn{3}{c}{\textbf{25\%}} & \multicolumn{3}{c}{\textbf{50\%}} & \multicolumn{3}{c}{\textbf{100\%}}\\
\cmidrule(lr){3-5}\cmidrule(lr){6-8}\cmidrule(lr){9-11}\cmidrule(lr){12-14}\cmidrule(lr){15-17}\cmidrule(lr){18-20}
 & & B-ACC & Kappa & F1 & B-ACC & Kappa & F1 & B-ACC & Kappa & F1 & B-ACC & Kappa & F1 & B-ACC & Kappa & F1 & B-ACC & Kappa & F1\\
\midrule
\endhead
\endfoot
\bottomrule
\endlastfoot
\multicolumn{20}{c}{\parbox[t]{10.4cm}{\centering \adjustbox{max width=\linewidth}{\strut \textbf{EEGMat: Task 1}}}}\\
\midrule
\multicolumn{20}{c}{\textbf{Baseline models}} \\
\midrule
\cellcolor{AppBaselineCell}ATCNet & Baseline & 42.31\% & \mbox{-17.65\%} & 40.74\% & 50.00\% & 0.00\% & 44.59\% & 51.84\% & 5.35\% & 48.12\% & 50.34\% & 0.98\% & 44.59\% & 49.84\% & \mbox{-0.48\%} & 43.35\% & 49.92\% & \mbox{-0.24\%} & 42.96\% \\
\addlinespace[3.0pt]
\cellcolor{AppBaselineCell}DeepConvNet & Baseline & 53.85\% & 3.03\% & 23.81\% & 57.95\% & 18.54\% & 58.86\% & 40.48\% & \mbox{-21.86\%} & 38.35\% & 48.08\% & \mbox{-5.27\%} & 43.23\% & 48.36\% & \mbox{-4.55\%} & 43.93\% & 50.00\% & 0.00\% & 43.00\% \\
\addlinespace[3.0pt]
\cellcolor{AppBaselineCell}EEGConformer & Baseline & 42.31\% & \mbox{-17.65\%} & 40.74\% & 50.00\% & 0.00\% & 44.59\% & 50.00\% & 0.00\% & 43.45\% & 50.00\% & 0.00\% & 42.94\% & 65.50\% & 31.58\% & 65.78\% & 65.46\% & 26.03\% & 62.21\% \\
\addlinespace[3.0pt]
\cellcolor{AppBaselineCell}EEGNet & Baseline & 50.00\% & 0.00\% & 44.83\% & 50.00\% & 0.00\% & 44.59\% & 50.00\% & 0.00\% & 43.45\% & 50.00\% & 0.00\% & 42.94\% & 50.00\% & 0.00\% & 43.42\% & 50.00\% & 0.00\% & 43.00\% \\
\addlinespace[3.0pt]
\cellcolor{AppBaselineCell}ShallowConvNet & Baseline & 50.00\% & 0.00\% & 44.83\% & 50.00\% & 0.00\% & 44.59\% & 50.00\% & 0.00\% & 43.45\% & 50.00\% & 0.00\% & 42.94\% & 63.35\% & 31.65\% & 65.20\% & 68.81\% & 41.71\% & 70.63\% \\
\addlinespace[2.2pt]
\midrule
\multicolumn{20}{c}{\textbf{EEG FMs}} \\
\end{longtable}
\endgroup

\begingroup
\setlength{\LTpre}{8pt}
\setlength{\LTpost}{8pt}
\fontsize{5.2}{5.95}\selectfont
\setlength{\tabcolsep}{0.05pt}
\renewcommand{\arraystretch}{0.94}
\begin{longtable}{@{}L{0.1040\linewidth}C{0.0750\linewidth}C{0.0431\linewidth}C{0.0431\linewidth}C{0.0431\linewidth}C{0.0431\linewidth}C{0.0431\linewidth}C{0.0431\linewidth}C{0.0431\linewidth}C{0.0431\linewidth}C{0.0431\linewidth}C{0.0431\linewidth}C{0.0431\linewidth}C{0.0431\linewidth}C{0.0431\linewidth}C{0.0431\linewidth}C{0.0431\linewidth}C{0.0431\linewidth}C{0.0431\linewidth}C{0.0431\linewidth}@{}}
\caption{Model--task performance by adaptation strategy and fine-tuning data budget for ERP-BCI and Kaggle-INRIA.}\label{tab:dense_matrix_table_iv_2}\\
\toprule
\multirow{2}{=}{\raggedright\textbf{Model}} & \multirow{2}{=}{\centering\textbf{Adaptation Strategies}} & \multicolumn{3}{c}{\textbf{2\%}} & \multicolumn{3}{c}{\textbf{5\%}} & \multicolumn{3}{c}{\textbf{10\%}} & \multicolumn{3}{c}{\textbf{25\%}} & \multicolumn{3}{c}{\textbf{50\%}} & \multicolumn{3}{c}{\textbf{100\%}}\\
\cmidrule(lr){3-5}\cmidrule(lr){6-8}\cmidrule(lr){9-11}\cmidrule(lr){12-14}\cmidrule(lr){15-17}\cmidrule(lr){18-20}
 & & B-ACC & Kappa & F1 & B-ACC & Kappa & F1 & B-ACC & Kappa & F1 & B-ACC & Kappa & F1 & B-ACC & Kappa & F1 & B-ACC & Kappa & F1\\
\midrule
\endfirsthead
\toprule
\multirow{2}{=}{\raggedright\textbf{Model}} & \multirow{2}{=}{\centering\textbf{Adaptation Strategies}} & \multicolumn{3}{c}{\textbf{2\%}} & \multicolumn{3}{c}{\textbf{5\%}} & \multicolumn{3}{c}{\textbf{10\%}} & \multicolumn{3}{c}{\textbf{25\%}} & \multicolumn{3}{c}{\textbf{50\%}} & \multicolumn{3}{c}{\textbf{100\%}}\\
\cmidrule(lr){3-5}\cmidrule(lr){6-8}\cmidrule(lr){9-11}\cmidrule(lr){12-14}\cmidrule(lr){15-17}\cmidrule(lr){18-20}
 & & B-ACC & Kappa & F1 & B-ACC & Kappa & F1 & B-ACC & Kappa & F1 & B-ACC & Kappa & F1 & B-ACC & Kappa & F1 & B-ACC & Kappa & F1\\
\midrule
\endhead
\endfoot
\bottomrule
\endlastfoot
\multicolumn{20}{c}{\parbox[t]{10.4cm}{\centering \adjustbox{max width=\linewidth}{\strut \textbf{ERP-BCI: Task 1}}}}\\
\midrule
\multicolumn{20}{c}{\textbf{Baseline models}} \\
\midrule
\cellcolor{AppBaselineCell}ATCNet & Baseline & - & - & - & - & - & - & - & - & - & - & - & - & - & - & - & - & - & - \\
\addlinespace[3.0pt]
\cellcolor{AppBaselineCell}DeepConvNet & Baseline & 45.98\% & 4.57\% & 22.60\% & 41.47\% & 1.01\% & 21.08\% & 50.00\% & 0.00\% & 45.78\% & 49.14\% & 0.22\% & 30.20\% & 49.98\% & \mbox{-0.04\%} & 30.19\% & 50.00\% & 0.00\% & 45.24\% \\
\addlinespace[3.0pt]
\cellcolor{AppBaselineCell}EEGConformer & Baseline & 50.86\% & 1.56\% & 50.68\% & 50.58\% & 1.33\% & 50.44\% & 52.60\% & 4.77\% & 52.28\% & 50.00\% & 0.00\% & 45.66\% & 50.00\% & 0.00\% & 45.30\% & 50.00\% & 0.00\% & 45.24\% \\
\addlinespace[3.0pt]
\cellcolor{AppBaselineCell}EEGNet & Baseline & 50.00\% & 0.00\% & 10.81\% & 50.00\% & 0.14\% & 8.54\% & 48.74\% & 0.01\% & 22.92\% & 56.50\% & 8.33\% & 25.10\% & 56.87\% & 12.17\% & 27.93\% & 56.63\% & 12.26\% & 28.00\% \\
\addlinespace[3.0pt]
\cellcolor{AppBaselineCell}ShallowConvNet & Baseline & 50.00\% & 0.28\% & 7.34\% & 52.59\% & 7.12\% & 25.22\% & 48.14\% & \mbox{-1.51\%} & 22.90\% & 47.95\% & 1.13\% & 23.78\% & 48.05\% & 1.80\% & 24.46\% & 48.23\% & 2.22\% & 23.84\% \\
\addlinespace[2.2pt]
\midrule
\multicolumn{20}{c}{\textbf{EEG FMs}} \\
\end{longtable}
\endgroup

\begingroup
\setlength{\LTpre}{8pt}
\setlength{\LTpost}{8pt}
\fontsize{5.2}{5.95}\selectfont
\setlength{\tabcolsep}{0.05pt}
\renewcommand{\arraystretch}{0.94}
\begin{longtable}{@{}L{0.1040\linewidth}C{0.0750\linewidth}C{0.0431\linewidth}C{0.0431\linewidth}C{0.0431\linewidth}C{0.0431\linewidth}C{0.0431\linewidth}C{0.0431\linewidth}C{0.0431\linewidth}C{0.0431\linewidth}C{0.0431\linewidth}C{0.0431\linewidth}C{0.0431\linewidth}C{0.0431\linewidth}C{0.0431\linewidth}C{0.0431\linewidth}C{0.0431\linewidth}C{0.0431\linewidth}C{0.0431\linewidth}C{0.0431\linewidth}@{}}
\caption{Model--task performance by adaptation strategy and fine-tuning data budget for SEED and HGD.}\label{tab:dense_matrix_table_iv_3}\\
\toprule
\multirow{2}{=}{\raggedright\textbf{Model}} & \multirow{2}{=}{\centering\textbf{Adaptation Strategies}} & \multicolumn{3}{c}{\textbf{2\%}} & \multicolumn{3}{c}{\textbf{5\%}} & \multicolumn{3}{c}{\textbf{10\%}} & \multicolumn{3}{c}{\textbf{25\%}} & \multicolumn{3}{c}{\textbf{50\%}} & \multicolumn{3}{c}{\textbf{100\%}}\\
\cmidrule(lr){3-5}\cmidrule(lr){6-8}\cmidrule(lr){9-11}\cmidrule(lr){12-14}\cmidrule(lr){15-17}\cmidrule(lr){18-20}
 & & B-ACC & Kappa & F1 & B-ACC & Kappa & F1 & B-ACC & Kappa & F1 & B-ACC & Kappa & F1 & B-ACC & Kappa & F1 & B-ACC & Kappa & F1\\
\midrule
\endfirsthead
\toprule
\multirow{2}{=}{\raggedright\textbf{Model}} & \multirow{2}{=}{\centering\textbf{Adaptation Strategies}} & \multicolumn{3}{c}{\textbf{2\%}} & \multicolumn{3}{c}{\textbf{5\%}} & \multicolumn{3}{c}{\textbf{10\%}} & \multicolumn{3}{c}{\textbf{25\%}} & \multicolumn{3}{c}{\textbf{50\%}} & \multicolumn{3}{c}{\textbf{100\%}}\\
\cmidrule(lr){3-5}\cmidrule(lr){6-8}\cmidrule(lr){9-11}\cmidrule(lr){12-14}\cmidrule(lr){15-17}\cmidrule(lr){18-20}
 & & B-ACC & Kappa & F1 & B-ACC & Kappa & F1 & B-ACC & Kappa & F1 & B-ACC & Kappa & F1 & B-ACC & Kappa & F1 & B-ACC & Kappa & F1\\
\midrule
\endhead
\endfoot
\bottomrule
\endlastfoot
\multicolumn{20}{c}{\parbox[t]{10.4cm}{\centering \adjustbox{max width=\linewidth}{\strut \textbf{SEED: Task 1}}}}\\
\midrule
\multicolumn{20}{c}{\textbf{Baseline models}} \\
\midrule
\cellcolor{AppBaselineCell}ATCNet & Baseline & 33.03\% & \mbox{-0.96\%} & 30.60\% & 37.75\% & 6.67\% & 36.95\% & 42.24\% & 13.45\% & 42.04\% & 52.75\% & 29.11\% & 49.17\% & 53.44\% & 30.37\% & 50.96\% & 52.30\% & 28.80\% & 49.90\% \\
\addlinespace[3.0pt]
\cellcolor{AppBaselineCell}DeepConvNet & Baseline & 30.44\% & \mbox{-4.82\%} & 27.47\% & 38.39\% & 7.55\% & 29.39\% & 40.45\% & 10.90\% & 32.60\% & 41.74\% & 12.48\% & 32.89\% & 44.96\% & 17.52\% & 36.06\% & 46.37\% & 19.86\% & 36.59\% \\
\addlinespace[3.0pt]
\cellcolor{AppBaselineCell}EEGConformer & Baseline & 28.01\% & \mbox{-7.12\%} & 20.66\% & 35.45\% & 3.05\% & 34.30\% & 39.94\% & 9.84\% & 39.39\% & 49.34\% & 24.04\% & 46.36\% & 54.49\% & 31.95\% & 51.69\% & 53.29\% & 30.22\% & 47.48\% \\
\addlinespace[3.0pt]
\cellcolor{AppBaselineCell}EEGNet & Baseline & 36.76\% & 4.86\% & 29.58\% & 31.63\% & \mbox{-2.45\%} & 25.36\% & 30.48\% & \mbox{-4.25\%} & 23.88\% & 51.77\% & 28.12\% & 51.37\% & 51.30\% & 27.17\% & 49.00\% & 51.23\% & 27.19\% & 49.61\% \\
\addlinespace[3.0pt]
\cellcolor{AppBaselineCell}ShallowConvNet & Baseline & 40.56\% & 11.11\% & 39.86\% & 41.83\% & 12.61\% & 40.38\% & 51.82\% & 28.45\% & 50.86\% & 52.03\% & 28.56\% & 50.72\% & 53.83\% & 31.07\% & 51.72\% & 54.25\% & 31.86\% & 52.65\% \\
\addlinespace[2.2pt]
\midrule
\multicolumn{20}{c}{\textbf{EEG FMs}} \\
\end{longtable}
\endgroup

\begingroup
\setlength{\LTpre}{8pt}
\setlength{\LTpost}{8pt}
\fontsize{5.2}{5.95}\selectfont
\setlength{\tabcolsep}{0.05pt}
\renewcommand{\arraystretch}{0.94}
\begin{longtable}{@{}L{0.1040\linewidth}C{0.0750\linewidth}C{0.0431\linewidth}C{0.0431\linewidth}C{0.0431\linewidth}C{0.0431\linewidth}C{0.0431\linewidth}C{0.0431\linewidth}C{0.0431\linewidth}C{0.0431\linewidth}C{0.0431\linewidth}C{0.0431\linewidth}C{0.0431\linewidth}C{0.0431\linewidth}C{0.0431\linewidth}C{0.0431\linewidth}C{0.0431\linewidth}C{0.0431\linewidth}C{0.0431\linewidth}C{0.0431\linewidth}@{}}
\caption{Model--task performance by adaptation strategy and fine-tuning data budget for Meng2019 and OpenBMI.}\label{tab:dense_matrix_table_iv_4}\\
\toprule
\multirow{2}{=}{\raggedright\textbf{Model}} & \multirow{2}{=}{\centering\textbf{Adaptation Strategies}} & \multicolumn{3}{c}{\textbf{2\%}} & \multicolumn{3}{c}{\textbf{5\%}} & \multicolumn{3}{c}{\textbf{10\%}} & \multicolumn{3}{c}{\textbf{25\%}} & \multicolumn{3}{c}{\textbf{50\%}} & \multicolumn{3}{c}{\textbf{100\%}}\\
\cmidrule(lr){3-5}\cmidrule(lr){6-8}\cmidrule(lr){9-11}\cmidrule(lr){12-14}\cmidrule(lr){15-17}\cmidrule(lr){18-20}
 & & B-ACC & Kappa & F1 & B-ACC & Kappa & F1 & B-ACC & Kappa & F1 & B-ACC & Kappa & F1 & B-ACC & Kappa & F1 & B-ACC & Kappa & F1\\
\midrule
\endfirsthead
\toprule
\multirow{2}{=}{\raggedright\textbf{Model}} & \multirow{2}{=}{\centering\textbf{Adaptation Strategies}} & \multicolumn{3}{c}{\textbf{2\%}} & \multicolumn{3}{c}{\textbf{5\%}} & \multicolumn{3}{c}{\textbf{10\%}} & \multicolumn{3}{c}{\textbf{25\%}} & \multicolumn{3}{c}{\textbf{50\%}} & \multicolumn{3}{c}{\textbf{100\%}}\\
\cmidrule(lr){3-5}\cmidrule(lr){6-8}\cmidrule(lr){9-11}\cmidrule(lr){12-14}\cmidrule(lr){15-17}\cmidrule(lr){18-20}
 & & B-ACC & Kappa & F1 & B-ACC & Kappa & F1 & B-ACC & Kappa & F1 & B-ACC & Kappa & F1 & B-ACC & Kappa & F1 & B-ACC & Kappa & F1\\
\midrule
\endhead
\endfoot
\bottomrule
\endlastfoot
\multicolumn{20}{c}{\parbox[t]{10.4cm}{\centering \adjustbox{max width=\linewidth}{\strut \textbf{Meng2019: Task 2}}}}\\
\midrule
\multicolumn{20}{c}{\textbf{Baseline models}} \\
\midrule
\cellcolor{AppBaselineCell}ATCNet & Baseline & 40.06\% & 12.62\% & 39.51\% & 39.79\% & 11.75\% & 39.16\% & 45.13\% & 20.11\% & 44.90\% & 45.64\% & 21.81\% & 45.46\% & 47.67\% & 25.65\% & 47.25\% & 49.70\% & 29.05\% & 49.66\% \\
\addlinespace[3.0pt]
\cellcolor{AppBaselineCell}DeepConvNet & Baseline & 39.29\% & 7.43\% & 26.67\% & 33.13\% & 0.37\% & 25.27\% & 33.12\% & 0.68\% & 27.07\% & 34.49\% & 2.87\% & 26.57\% & 37.99\% & 7.61\% & 32.78\% & 41.02\% & 11.39\% & 35.96\% \\
\addlinespace[3.0pt]
\cellcolor{AppBaselineCell}EEGConformer & Baseline & 44.15\% & 18.98\% & 44.87\% & 42.75\% & 17.64\% & 42.69\% & 44.37\% & 19.32\% & 44.30\% & 42.30\% & 16.91\% & 41.98\% & 45.61\% & 22.25\% & 45.18\% & 49.43\% & 27.99\% & 49.31\% \\
\addlinespace[3.0pt]
\cellcolor{AppBaselineCell}EEGNet & Baseline & 40.73\% & 11.60\% & 30.01\% & 32.91\% & 1.11\% & 32.74\% & 38.29\% & 9.26\% & 37.39\% & 44.75\% & 20.61\% & 44.56\% & 46.24\% & 23.00\% & 45.93\% & 47.30\% & 25.17\% & 47.19\% \\
\addlinespace[3.0pt]
\cellcolor{AppBaselineCell}ShallowConvNet & Baseline & 43.28\% & 14.45\% & 37.92\% & 42.11\% & 14.47\% & 41.48\% & 45.89\% & 20.05\% & 45.65\% & 45.60\% & 20.95\% & 45.45\% & 51.09\% & 29.91\% & 51.21\% & 54.08\% & 35.44\% & 54.47\% \\
\addlinespace[2.2pt]
\midrule
\multicolumn{20}{c}{\textbf{EEG FMs}} \\
\end{longtable}
\endgroup

\begingroup
\setlength{\LTpre}{8pt}
\setlength{\LTpost}{8pt}
\fontsize{5.2}{5.95}\selectfont
\setlength{\tabcolsep}{0.05pt}
\renewcommand{\arraystretch}{0.94}
\begin{longtable}{@{}L{0.1040\linewidth}C{0.0750\linewidth}C{0.0431\linewidth}C{0.0431\linewidth}C{0.0431\linewidth}C{0.0431\linewidth}C{0.0431\linewidth}C{0.0431\linewidth}C{0.0431\linewidth}C{0.0431\linewidth}C{0.0431\linewidth}C{0.0431\linewidth}C{0.0431\linewidth}C{0.0431\linewidth}C{0.0431\linewidth}C{0.0431\linewidth}C{0.0431\linewidth}C{0.0431\linewidth}C{0.0431\linewidth}C{0.0431\linewidth}@{}}
\caption{Model--task performance by adaptation strategy and fine-tuning data budget for PhysioNet-MI.}\label{tab:dense_matrix_table_iv_5}\label{tab:dense_matrix_table_iv_last}\\
\toprule
 &  & \multicolumn{18}{c}{\parbox[t]{10.4cm}{\centering \adjustbox{max width=\linewidth}{\strut \textbf{PhysioNet-MI: Task 3}}}}\\
\cmidrule(lr){3-20}
\textbf{Model} & \textbf{Adaptation Strategies} & \multicolumn{3}{c}{\textbf{2\%}} & \multicolumn{3}{c}{\textbf{5\%}} & \multicolumn{3}{c}{\textbf{10\%}} & \multicolumn{3}{c}{\textbf{25\%}} & \multicolumn{3}{c}{\textbf{50\%}} & \multicolumn{3}{c}{\textbf{100\%}}\\
\cmidrule(lr){3-5}\cmidrule(lr){6-8}\cmidrule(lr){9-11}\cmidrule(lr){12-14}\cmidrule(lr){15-17}\cmidrule(lr){18-20}
 & & B-ACC & Kappa & F1 & B-ACC & Kappa & F1 & B-ACC & Kappa & F1 & B-ACC & Kappa & F1 & B-ACC & Kappa & F1 & B-ACC & Kappa & F1\\
\midrule
\endfirsthead
\toprule
 &  & \multicolumn{18}{c}{\parbox[t]{10.4cm}{\centering \adjustbox{max width=\linewidth}{\strut \textbf{PhysioNet-MI: Task 3}}}}\\
\cmidrule(lr){3-20}
\textbf{Model} & \textbf{Adaptation Strategies} & \multicolumn{3}{c}{\textbf{2\%}} & \multicolumn{3}{c}{\textbf{5\%}} & \multicolumn{3}{c}{\textbf{10\%}} & \multicolumn{3}{c}{\textbf{25\%}} & \multicolumn{3}{c}{\textbf{50\%}} & \multicolumn{3}{c}{\textbf{100\%}}\\
\cmidrule(lr){3-5}\cmidrule(lr){6-8}\cmidrule(lr){9-11}\cmidrule(lr){12-14}\cmidrule(lr){15-17}\cmidrule(lr){18-20}
 & & B-ACC & Kappa & F1 & B-ACC & Kappa & F1 & B-ACC & Kappa & F1 & B-ACC & Kappa & F1 & B-ACC & Kappa & F1 & B-ACC & Kappa & F1\\
\midrule
\endhead
\endfoot
\bottomrule
\endlastfoot
\multicolumn{20}{c}{\textbf{Baseline models}} \\
\midrule
\cellcolor{AppBaselineCell}ATCNet & Baseline & 50.00\% & 0.00\% & 43.48\% & 46.91\% & \mbox{-4.12\%} & 28.95\% & 44.93\% & 4.33\% & 20.65\% & 51.47\% & 4.83\% & 24.59\% & 58.73\% & 21.47\% & 59.06\% & 62.48\% & 27.43\% & 63.47\% \\
\addlinespace[3.0pt]
\cellcolor{AppBaselineCell}DeepConvNet & Baseline & 3.33\% & \mbox{-4.57\%} & 3.12\% & 50.62\% & 1.49\% & 50.08\% & 53.22\% & 6.99\% & 53.25\% & 59.64\% & 21.31\% & 60.44\% & 58.69\% & 20.79\% & 59.13\% & 64.92\% & 30.15\% & 65.07\% \\
\addlinespace[3.0pt]
\cellcolor{AppBaselineCell}EEGConformer & Baseline & 55.56\% & 16.13\% & 54.12\% & 57.72\% & 16.50\% & 58.18\% & 49.66\% & \mbox{-1.00\%} & 42.61\% & 57.27\% & 16.63\% & 57.90\% & 60.51\% & 19.58\% & 59.60\% & 64.41\% & 27.14\% & 63.47\% \\
\addlinespace[3.0pt]
\cellcolor{AppBaselineCell}EEGNet & Baseline & 50.00\% & 0.00\% & 43.48\% & 49.38\% & 3.24\% & 29.96\% & 48.65\% & 1.78\% & 28.32\% & 52.60\% & 6.75\% & 51.54\% & 55.33\% & 13.21\% & 54.59\% & 53.54\% & 9.33\% & 51.90\% \\
\addlinespace[3.0pt]
\cellcolor{AppBaselineCell}ShallowConvNet & Baseline & 35.00\% & \mbox{-5.85\%} & 12.83\% & 38.27\% & 1.74\% & 21.37\% & 48.19\% & 10.53\% & 28.98\% & 52.39\% & 6.53\% & 50.56\% & 57.09\% & 17.82\% & 37.87\% & 64.02\% & 29.16\% & 43.03\% \\
\addlinespace[2.2pt]
\midrule
\multicolumn{20}{c}{\textbf{EEG FMs}} \\
\midrule
\cellcolor{AppFMCell} & \mbox{Linear Probe} & 50.00\% & 0.00\% & 43.48\% & 58.33\% & 14.48\% & 56.92\% & 53.23\% & 7.22\% & 53.16\% & 61.61\% & 20.94\% & 60.24\% & 60.81\% & 19.62\% & 59.44\% & 62.18\% & 20.96\% & 59.79\% \\
\end{longtable}
\endgroup

\FloatBarrier

\subsubsection{Data Budget and Adaptation Strategy Curves}
\label{app:fine_tuning_data_budget_curves}

Figures~\ref{fig:app-data-budget-curves-hgd}--\ref{fig:app-data-budget-curves-seed} show the data-budget trajectories for the evaluated datasets.

\begingroup
\setlength{\intextsep}{0pt}
\setlength{\abovecaptionskip}{2pt}
\setlength{\belowcaptionskip}{0pt}
\begin{figure}[H]
  \centering
  \includegraphics[width=\linewidth,height=0.82\textheight,keepaspectratio]{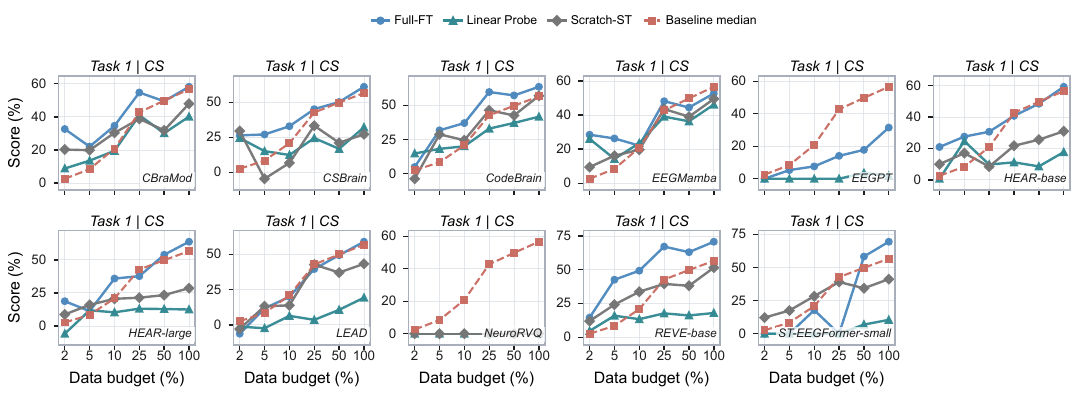}
  \caption{Data budget and adaptation strategy curves for HGD.}
  \label{fig:app-data-budget-curves-hgd}
\end{figure}

\begin{figure}[H]
  \centering
  \includegraphics[width=\linewidth,height=0.82\textheight,keepaspectratio]{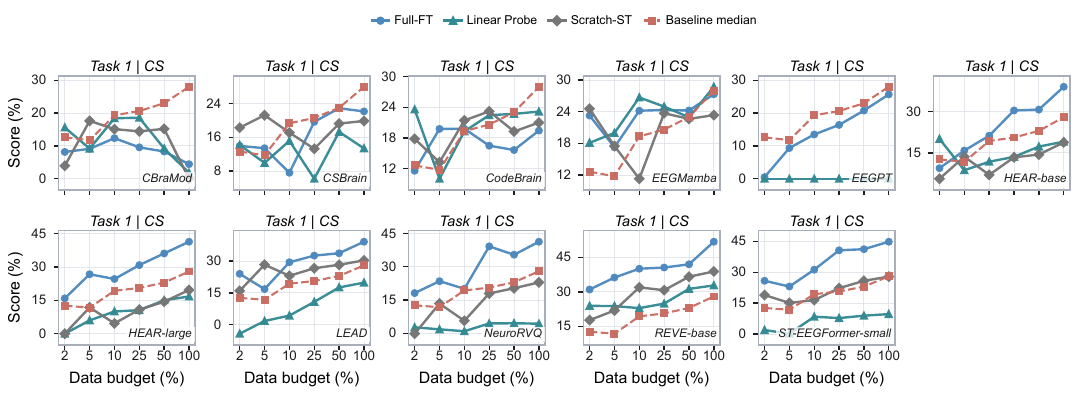}
  \caption{Data budget and adaptation strategy curves for Meng2019.}
  \label{fig:app-data-budget-curves-meng2019}
\end{figure}

\begin{figure}[H]
  \centering
  \includegraphics[width=\linewidth,height=0.82\textheight,keepaspectratio]{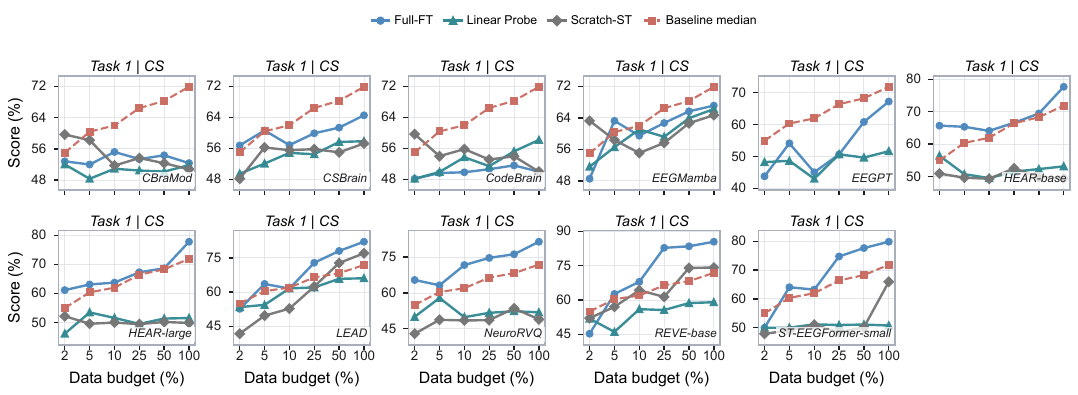}
  \caption{Data budget and adaptation strategy curves for OpenBMI.}
  \label{fig:app-data-budget-curves-openbmi}
\end{figure}

\begin{figure}[H]
  \centering
  \includegraphics[width=\linewidth,height=0.82\textheight,keepaspectratio]{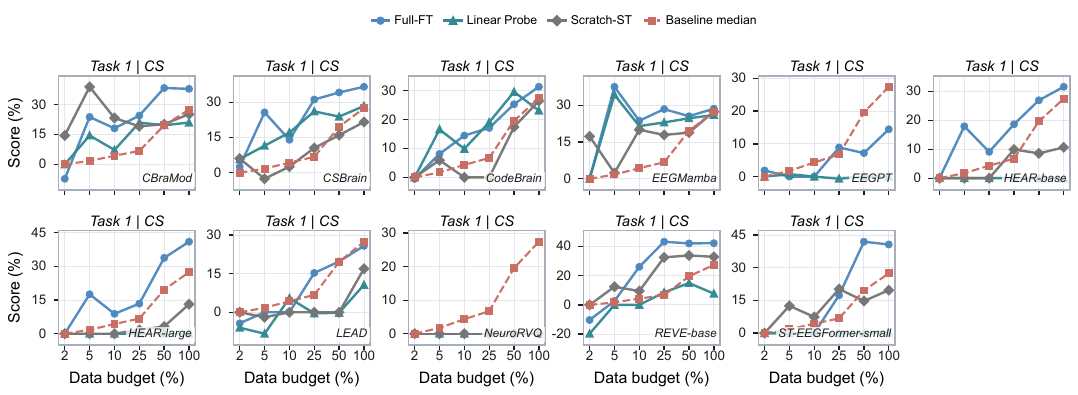}
  \caption{Data budget and adaptation strategy curves for PhysioNet-MI.}
  \label{fig:app-data-budget-curves-physionet-mi}
\end{figure}

\begin{figure}[H]
  \centering
  \includegraphics[width=\linewidth,height=0.82\textheight,keepaspectratio]{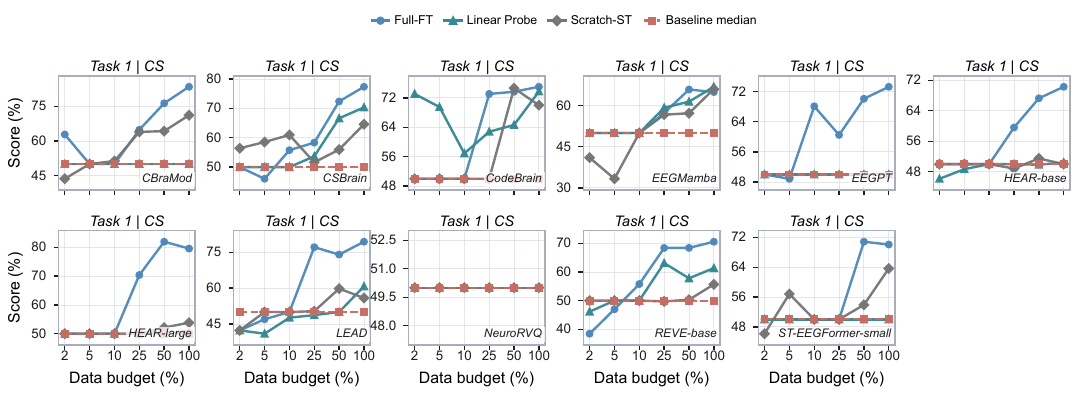}
  \caption{Data budget and adaptation strategy curves for EEGMat.}
  \label{fig:app-data-budget-curves-eegmat}
\end{figure}

\begin{figure}[H]
  \centering
  \includegraphics[width=\linewidth,height=0.82\textheight,keepaspectratio]{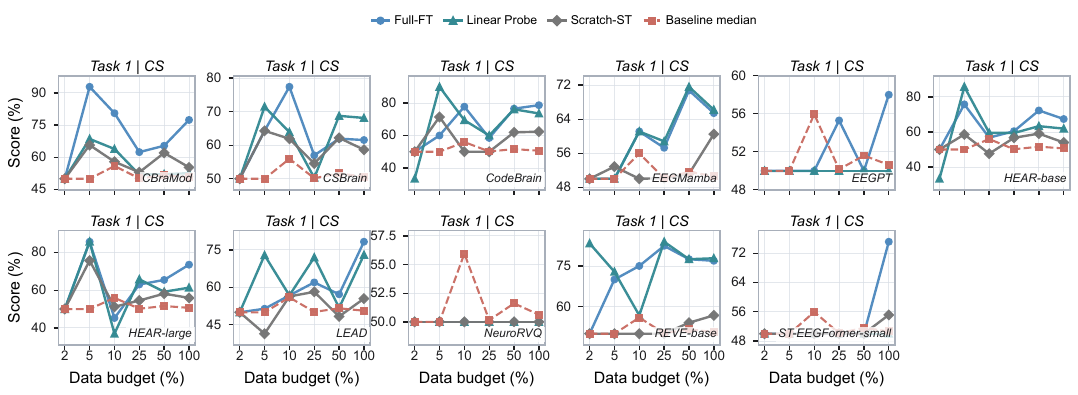}
  \caption{Data budget and adaptation strategy curves for TUSL.}
  \label{fig:app-data-budget-curves-tusl}
\end{figure}

\begin{figure}[H]
  \centering
  \includegraphics[width=\linewidth,height=0.82\textheight,keepaspectratio]{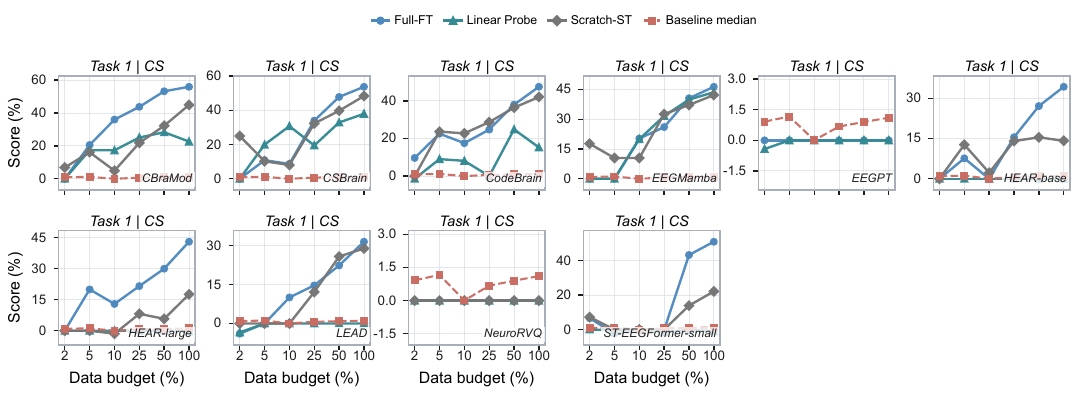}
  \caption{Data budget and adaptation strategy curves for ERP-BCI.}
  \label{fig:app-data-budget-curves-erp-bci}
\end{figure}

\begin{figure}[H]
  \centering
  \includegraphics[width=\linewidth,height=0.82\textheight,keepaspectratio]{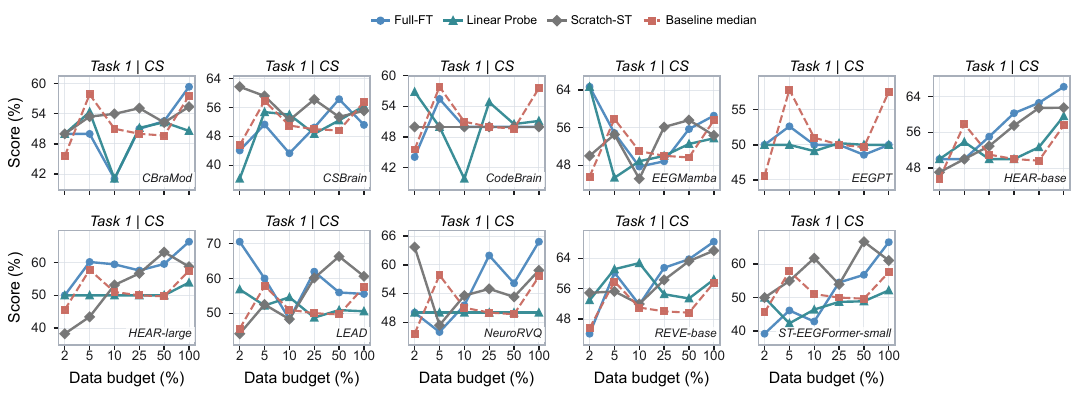}
  \caption{Data budget and adaptation strategy curves for Kaggle-INRIA.}
  \label{fig:app-data-budget-curves-kaggle-inria}
\end{figure}

\begin{figure}[H]
  \centering
  \includegraphics[width=\linewidth,height=0.82\textheight,keepaspectratio]{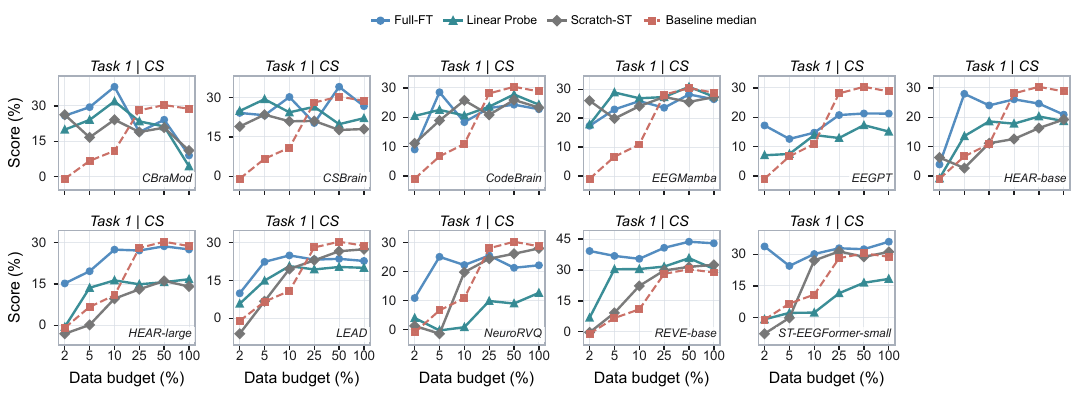}
  \caption{Data budget and adaptation strategy curves for SEED.}
  \label{fig:app-data-budget-curves-seed}
\end{figure}

\FloatBarrier
\endgroup

\subsubsection{Data Budget Curves by Task and Model}
\label{app:data_budget_extra_detail_figures}

Figures~\ref{fig:app-data-budget-task-strategy-curves-01}--\ref{fig:app-data-budget-model-strategy-curves-01} present the learning curves across data budgets and adaptation strategy by task and model.

\begingroup
\setlength{\intextsep}{0pt}
\setlength{\abovecaptionskip}{2pt}
\begin{figure}[H]
  \centering
  \includegraphics[width=\linewidth]{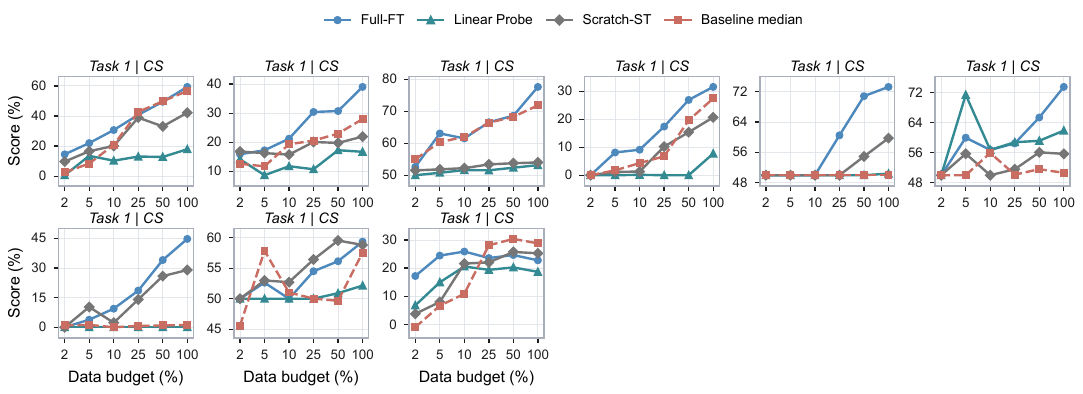}
  \caption{Task-level trajectories by adaptation strategy and fine-tuning data budget.}
  \label{fig:app-data-budget-task-strategy-curves-01}
\end{figure}

\begin{figure}[H]
  \centering
  \includegraphics[width=\linewidth]{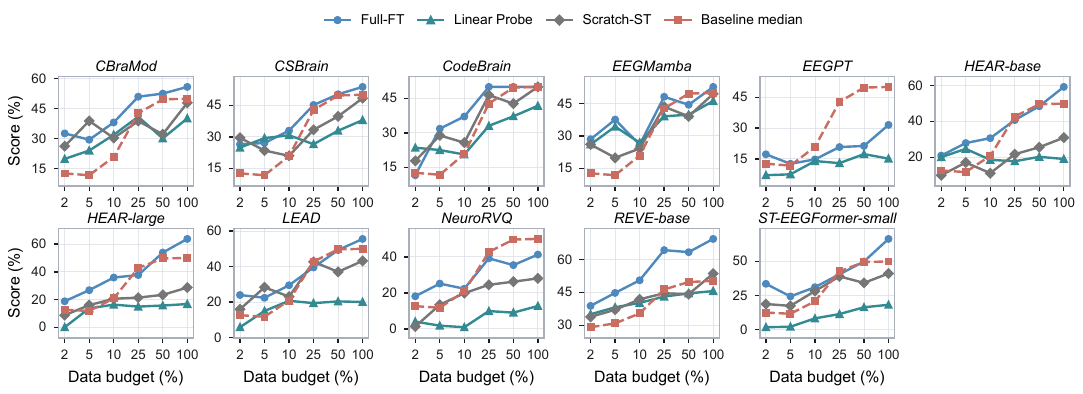}
  \caption{Model-level trajectories by adaptation strategy and fine-tuning data budget.}
  \label{fig:app-data-budget-model-strategy-curves-01}
\end{figure}

\FloatBarrier
\endgroup

\endgroup
\FloatBarrier

\subsection{Protocol III: The Scaling of Model Size}
\label{app:model_scaling_results}

Tables~\ref{tab:protocol_iii_model_scaling_results}--\ref{tab:protocol_iii_model_scaling_results_last} report the Linear Probe and Full-FT results for all the EEG FM variants evaluated in Protocol III. Each table contains three tasks with WS and CS results in B-ACC, Cohen's kappa, and macro-F1.

\begingroup
\definecolor{AppScalingHEAR}{HTML}{F4E6EF}
\definecolor{AppScalingLUNA}{HTML}{E3F3F1}
\definecolor{AppScalingREVE}{HTML}{E7EEF8}
\definecolor{AppScalingST}{HTML}{EEE9F7}
\input{appendix_files/tex/app_tab_protocol_iii_model_scaling_results.tex}
\endgroup

\FloatBarrier

\subsection{Protocol V: Robustness to Channel Configuration}
\label{app:input_space_retention_results}

In this section, we report the channel robustness analysis results across the evaluated models and the channel configurations used in the experiments.

\subsubsection{Channel Configurations}

Tables~\ref{tab:channel_subset_inventory} and~\ref{tab:model_specific_channel_inputs} provide definitions of channel configurations across datasets, tasks, and models.

\begingroup
\setlength{\LTpre}{2pt}
\setlength{\LTpost}{0pt}
\fontsize{6.30}{6.90}\selectfont
\setlength{\tabcolsep}{1.25pt}
\renewcommand{\arraystretch}{1.02}
\begin{longtable}{@{}C{0.07779\linewidth}C{0.06254\linewidth}C{0.06559\linewidth}L{0.77409\linewidth}@{}}
\caption{Channel configurations across evaluated datasets and tasks.}\label{tab:channel_subset_inventory}\\
\toprule
Subset & Ratio & Count & Electrodes\\
\midrule
\endfirsthead
\toprule
Subset & Ratio & Count & Electrodes\\
\midrule
\endhead
\endfoot
\bottomrule
\endlastfoot
\multicolumn{4}{c}{\textbf{BCIC-IV-2B: Task 1}} \\
\midrule
s01 & 33.33\% & 1/3 & C3 \\
\addlinespace[2.4pt]
s02 & 33.33\% & 1/3 & Cz \\
\addlinespace[2.4pt]
s03 & 33.33\% & 1/3 & C4 \\
\addlinespace[2.4pt]
s04 & 66.67\% & 2/3 & C3, Cz \\
\addlinespace[2.4pt]
s05 & 66.67\% & 2/3 & Cz, C4 \\
\addlinespace[2.4pt]
s06 & 66.67\% & 2/3 & C3, C4 \\
\addlinespace[2.4pt]
s07 & 100.00\% & 3/3 & C3, Cz, C4 \\
\end{longtable}
\endgroup

\par\bigskip
\Needspace{0.56\textheight}

\begingroup
\setlength{\LTpre}{2pt}
\setlength{\LTpost}{0pt}
\fontsize{6.25}{6.85}\selectfont
\setlength{\tabcolsep}{1.25pt}
\renewcommand{\arraystretch}{0.96}
\begin{longtable}{@{}C{0.05428\linewidth}C{0.10705\linewidth}C{0.09800\linewidth}L{0.72068\linewidth}@{}}
\caption{Model-specific channel inputs.}\label{tab:model_specific_channel_inputs}\\
\toprule
Subset & Subset number & Input number & Input electrodes\\
\midrule
\endfirsthead
\toprule
Subset & Subset number & Input number & Input electrodes\\
\midrule
\endhead
\endfoot
\bottomrule
\endlastfoot
\multicolumn{4}{c}{\textbf{CBraMod} \textit{(Channel-constrained)}} \\*
\midrule
\multicolumn{4}{@{}l@{}}{\hspace{0.5em}\textbf{BCIC-IV-2B}: Task 1} \\*
\cmidrule(lr){1-4}
s01 & 1/3 & 1 & C3 \\
\addlinespace[1.5pt]
s02 & 1/3 & 1 & Cz \\
\addlinespace[1.5pt]
s03 & 1/3 & 1 & C4 \\
\addlinespace[1.5pt]
s04 & 2/3 & 2 & C3, Cz \\
\addlinespace[1.5pt]
s05 & 2/3 & 2 & Cz, C4 \\
\addlinespace[1.5pt]
s06 & 2/3 & 2 & C3, C4 \\
\end{longtable}
\endgroup

\par\bigskip

\subsubsection{Channel Robustness Results of Evaluated EEG FMs and Supervised Baselines}

\begingroup
\setlength{\LTpre}{8pt}
\setlength{\LTpost}{8pt}
\fontsize{4.35}{4.95}\selectfont
\setlength{\tabcolsep}{0.10pt}
\renewcommand{\arraystretch}{0.90}
\begin{longtable}{@{}C{0.03577\linewidth}C{0.03147\linewidth}C{0.03147\linewidth}C{0.03147\linewidth}C{0.03147\linewidth}C{0.03147\linewidth}C{0.03147\linewidth}C{0.03147\linewidth}C{0.03147\linewidth}C{0.03147\linewidth}C{0.03147\linewidth}C{0.03147\linewidth}C{0.03147\linewidth}C{0.03147\linewidth}C{0.03147\linewidth}C{0.03147\linewidth}C{0.03147\linewidth}C{0.03147\linewidth}C{0.03147\linewidth}C{0.03147\linewidth}C{0.03147\linewidth}C{0.03147\linewidth}C{0.03147\linewidth}C{0.03147\linewidth}C{0.03147\linewidth}C{0.03147\linewidth}C{0.03147\linewidth}C{0.03147\linewidth}C{0.03147\linewidth}C{0.03147\linewidth}C{0.03147\linewidth}@{}}
\caption{Channel robustness results for ERP-BCI, Kaggle-INRIA, and BCIC-IV-2B.}\label{tab:dense_matrix_table_v}\\
\toprule
\multicolumn{31}{l}{\textit{Best values are bolded within each task metric.}}\\
\midrule
\endfirsthead
\toprule
\textbf{Subset} & \multicolumn{3}{c}{\adjustbox{max width=1.26cm}{\strut \textbf{CBraMod}}} & \multicolumn{3}{c}{\adjustbox{max width=1.26cm}{\strut \textbf{CodeBrain}}} & \multicolumn{3}{c}{\adjustbox{max width=1.26cm}{\strut \textbf{CSBrain}}} & \multicolumn{3}{c}{\adjustbox{max width=1.26cm}{\strut \textbf{EEGPT}}} & \multicolumn{3}{c}{\adjustbox{max width=1.26cm}{\strut \textbf{NeuroRVQ}}} & \multicolumn{3}{c}{\adjustbox{max width=1.26cm}{\strut \textbf{REVE-base}}} & \multicolumn{3}{c}{\adjustbox{max width=1.26cm}{\strut \textbf{ST-EEGFormer-small}}} & \multicolumn{3}{c}{\adjustbox{max width=1.26cm}{\strut \textbf{LEAD}}} & \multicolumn{3}{c}{\adjustbox{max width=1.26cm}{\strut \textbf{HEAR-base}}} & \multicolumn{3}{c}{\adjustbox{max width=1.26cm}{\strut \textbf{HEAR-large}}}\\
\cmidrule(lr){2-4}\cmidrule(lr){5-7}\cmidrule(lr){8-10}\cmidrule(lr){11-13}\cmidrule(lr){14-16}\cmidrule(lr){17-19}\cmidrule(lr){20-22}\cmidrule(lr){23-25}\cmidrule(lr){26-28}\cmidrule(lr){29-31}
 & B-ACC & Kap. & F1 & B-ACC & Kap. & F1 & B-ACC & Kap. & F1 & B-ACC & Kap. & F1 & B-ACC & Kap. & F1 & B-ACC & Kap. & F1 & B-ACC & Kap. & F1 & B-ACC & Kap. & F1 & B-ACC & Kap. & F1 & B-ACC & Kap. & F1\\
\midrule
\endhead
\endfoot
\bottomrule
\endlastfoot
\multicolumn{31}{c}{\adjustbox{max width=\linewidth}{\strut \textbf{ERP-BCI: Task 1 (CS)}}} \\*
\cmidrule(lr){1-31}
\textbf{Subset} & \multicolumn{3}{c}{\adjustbox{max width=1.26cm}{\strut \textbf{CBraMod}}} & \multicolumn{3}{c}{\adjustbox{max width=1.26cm}{\strut \textbf{CodeBrain}}} & \multicolumn{3}{c}{\adjustbox{max width=1.26cm}{\strut \textbf{CSBrain}}} & \multicolumn{3}{c}{\adjustbox{max width=1.26cm}{\strut \textbf{EEGPT}}} & \multicolumn{3}{c}{\adjustbox{max width=1.26cm}{\strut \textbf{NeuroRVQ}}} & \multicolumn{3}{c}{\adjustbox{max width=1.26cm}{\strut \textbf{REVE-base}}} & \multicolumn{3}{c}{\adjustbox{max width=1.26cm}{\strut \textbf{ST-EEGFormer-small}}} & \multicolumn{3}{c}{\adjustbox{max width=1.26cm}{\strut \textbf{LEAD}}} & \multicolumn{3}{c}{\adjustbox{max width=1.26cm}{\strut \textbf{HEAR-base}}} & \multicolumn{3}{c}{\adjustbox{max width=1.26cm}{\strut \textbf{HEAR-large}}}\\
\cmidrule(lr){2-4}\cmidrule(lr){5-7}\cmidrule(lr){8-10}\cmidrule(lr){11-13}\cmidrule(lr){14-16}\cmidrule(lr){17-19}\cmidrule(lr){20-22}\cmidrule(lr){23-25}\cmidrule(lr){26-28}\cmidrule(lr){29-31}
 & B-ACC & Kap. & F1 & B-ACC & Kap. & F1 & B-ACC & Kap. & F1 & B-ACC & Kap. & F1 & B-ACC & Kap. & F1 & B-ACC & Kap. & F1 & B-ACC & Kap. & F1 & B-ACC & Kap. & F1 & B-ACC & Kap. & F1 & B-ACC & Kap. & F1\\
\end{longtable}
\endgroup

\begingroup
\setlength{\LTpre}{8pt}
\setlength{\LTpost}{8pt}
\fontsize{4.35}{4.95}\selectfont
\setlength{\tabcolsep}{0.10pt}
\renewcommand{\arraystretch}{0.90}
\begin{longtable}{@{}C{0.03577\linewidth}C{0.03147\linewidth}C{0.03147\linewidth}C{0.03147\linewidth}C{0.03147\linewidth}C{0.03147\linewidth}C{0.03147\linewidth}C{0.03147\linewidth}C{0.03147\linewidth}C{0.03147\linewidth}C{0.03147\linewidth}C{0.03147\linewidth}C{0.03147\linewidth}C{0.03147\linewidth}C{0.03147\linewidth}C{0.03147\linewidth}C{0.03147\linewidth}C{0.03147\linewidth}C{0.03147\linewidth}C{0.03147\linewidth}C{0.03147\linewidth}C{0.03147\linewidth}C{0.03147\linewidth}C{0.03147\linewidth}C{0.03147\linewidth}C{0.03147\linewidth}C{0.03147\linewidth}C{0.03147\linewidth}C{0.03147\linewidth}C{0.03147\linewidth}C{0.03147\linewidth}@{}}
\caption{Channel robustness results for HGD, Meng2019, and OpenBMI.}\label{tab:dense_matrix_table_v_2}\\
\toprule
\multicolumn{31}{l}{\textit{Best values are bolded within each task metric.}}\\
\midrule
\endfirsthead
\toprule
\textbf{Subset} & \multicolumn{3}{c}{\adjustbox{max width=1.26cm}{\strut \textbf{CBraMod}}} & \multicolumn{3}{c}{\adjustbox{max width=1.26cm}{\strut \textbf{CodeBrain}}} & \multicolumn{3}{c}{\adjustbox{max width=1.26cm}{\strut \textbf{CSBrain}}} & \multicolumn{3}{c}{\adjustbox{max width=1.26cm}{\strut \textbf{EEGPT}}} & \multicolumn{3}{c}{\adjustbox{max width=1.26cm}{\strut \textbf{NeuroRVQ}}} & \multicolumn{3}{c}{\adjustbox{max width=1.26cm}{\strut \textbf{REVE-base}}} & \multicolumn{3}{c}{\adjustbox{max width=1.26cm}{\strut \textbf{ST-EEGFormer-small}}} & \multicolumn{3}{c}{\adjustbox{max width=1.26cm}{\strut \textbf{LEAD}}} & \multicolumn{3}{c}{\adjustbox{max width=1.26cm}{\strut \textbf{HEAR-base}}} & \multicolumn{3}{c}{\adjustbox{max width=1.26cm}{\strut \textbf{HEAR-large}}}\\
\cmidrule(lr){2-4}\cmidrule(lr){5-7}\cmidrule(lr){8-10}\cmidrule(lr){11-13}\cmidrule(lr){14-16}\cmidrule(lr){17-19}\cmidrule(lr){20-22}\cmidrule(lr){23-25}\cmidrule(lr){26-28}\cmidrule(lr){29-31}
 & B-ACC & Kap. & F1 & B-ACC & Kap. & F1 & B-ACC & Kap. & F1 & B-ACC & Kap. & F1 & B-ACC & Kap. & F1 & B-ACC & Kap. & F1 & B-ACC & Kap. & F1 & B-ACC & Kap. & F1 & B-ACC & Kap. & F1 & B-ACC & Kap. & F1\\
\midrule
\endhead
\endfoot
\bottomrule
\endlastfoot
\multicolumn{31}{c}{\adjustbox{max width=\linewidth}{\strut \textbf{HGD: Task 1 (CS)}}} \\*
\cmidrule(lr){1-31}
\textbf{Subset} & \multicolumn{3}{c}{\adjustbox{max width=1.26cm}{\strut \textbf{CBraMod}}} & \multicolumn{3}{c}{\adjustbox{max width=1.26cm}{\strut \textbf{CodeBrain}}} & \multicolumn{3}{c}{\adjustbox{max width=1.26cm}{\strut \textbf{CSBrain}}} & \multicolumn{3}{c}{\adjustbox{max width=1.26cm}{\strut \textbf{EEGPT}}} & \multicolumn{3}{c}{\adjustbox{max width=1.26cm}{\strut \textbf{NeuroRVQ}}} & \multicolumn{3}{c}{\adjustbox{max width=1.26cm}{\strut \textbf{REVE-base}}} & \multicolumn{3}{c}{\adjustbox{max width=1.26cm}{\strut \textbf{ST-EEGFormer-small}}} & \multicolumn{3}{c}{\adjustbox{max width=1.26cm}{\strut \textbf{LEAD}}} & \multicolumn{3}{c}{\adjustbox{max width=1.26cm}{\strut \textbf{HEAR-base}}} & \multicolumn{3}{c}{\adjustbox{max width=1.26cm}{\strut \textbf{HEAR-large}}}\\
\cmidrule(lr){2-4}\cmidrule(lr){5-7}\cmidrule(lr){8-10}\cmidrule(lr){11-13}\cmidrule(lr){14-16}\cmidrule(lr){17-19}\cmidrule(lr){20-22}\cmidrule(lr){23-25}\cmidrule(lr){26-28}\cmidrule(lr){29-31}
 & B-ACC & Kap. & F1 & B-ACC & Kap. & F1 & B-ACC & Kap. & F1 & B-ACC & Kap. & F1 & B-ACC & Kap. & F1 & B-ACC & Kap. & F1 & B-ACC & Kap. & F1 & B-ACC & Kap. & F1 & B-ACC & Kap. & F1 & B-ACC & Kap. & F1\\
\end{longtable}
\endgroup

\begingroup
\setlength{\LTpre}{8pt}
\setlength{\LTpost}{8pt}
\fontsize{4.35}{4.95}\selectfont
\setlength{\tabcolsep}{0.10pt}
\renewcommand{\arraystretch}{0.90}
\begin{longtable}{@{}C{0.03577\linewidth}C{0.03147\linewidth}C{0.03147\linewidth}C{0.03147\linewidth}C{0.03147\linewidth}C{0.03147\linewidth}C{0.03147\linewidth}C{0.03147\linewidth}C{0.03147\linewidth}C{0.03147\linewidth}C{0.03147\linewidth}C{0.03147\linewidth}C{0.03147\linewidth}C{0.03147\linewidth}C{0.03147\linewidth}C{0.03147\linewidth}C{0.03147\linewidth}C{0.03147\linewidth}C{0.03147\linewidth}C{0.03147\linewidth}C{0.03147\linewidth}C{0.03147\linewidth}C{0.03147\linewidth}C{0.03147\linewidth}C{0.03147\linewidth}C{0.03147\linewidth}C{0.03147\linewidth}C{0.03147\linewidth}C{0.03147\linewidth}C{0.03147\linewidth}C{0.03147\linewidth}@{}}
\caption{Channel robustness results for PhysioNet-MI and SHU-MI.}\label{tab:dense_matrix_table_v_3}\label{tab:dense_matrix_table_v_last}\\
\toprule
\multicolumn{31}{l}{\textit{Best values are bolded within each task metric.}}\\
\midrule
\endfirsthead
\toprule
\textbf{Subset} & \multicolumn{3}{c}{\adjustbox{max width=1.26cm}{\strut \textbf{CBraMod}}} & \multicolumn{3}{c}{\adjustbox{max width=1.26cm}{\strut \textbf{CodeBrain}}} & \multicolumn{3}{c}{\adjustbox{max width=1.26cm}{\strut \textbf{CSBrain}}} & \multicolumn{3}{c}{\adjustbox{max width=1.26cm}{\strut \textbf{EEGPT}}} & \multicolumn{3}{c}{\adjustbox{max width=1.26cm}{\strut \textbf{NeuroRVQ}}} & \multicolumn{3}{c}{\adjustbox{max width=1.26cm}{\strut \textbf{REVE-base}}} & \multicolumn{3}{c}{\adjustbox{max width=1.26cm}{\strut \textbf{ST-EEGFormer-small}}} & \multicolumn{3}{c}{\adjustbox{max width=1.26cm}{\strut \textbf{LEAD}}} & \multicolumn{3}{c}{\adjustbox{max width=1.26cm}{\strut \textbf{HEAR-base}}} & \multicolumn{3}{c}{\adjustbox{max width=1.26cm}{\strut \textbf{HEAR-large}}}\\
\cmidrule(lr){2-4}\cmidrule(lr){5-7}\cmidrule(lr){8-10}\cmidrule(lr){11-13}\cmidrule(lr){14-16}\cmidrule(lr){17-19}\cmidrule(lr){20-22}\cmidrule(lr){23-25}\cmidrule(lr){26-28}\cmidrule(lr){29-31}
 & B-ACC & Kap. & F1 & B-ACC & Kap. & F1 & B-ACC & Kap. & F1 & B-ACC & Kap. & F1 & B-ACC & Kap. & F1 & B-ACC & Kap. & F1 & B-ACC & Kap. & F1 & B-ACC & Kap. & F1 & B-ACC & Kap. & F1 & B-ACC & Kap. & F1\\
\midrule
\endhead
\endfoot
\bottomrule
\endlastfoot
\multicolumn{31}{c}{\adjustbox{max width=\linewidth}{\strut \textbf{PhysioNet-MI: Task 3 (CS)}}} \\*
\cmidrule(lr){1-31}
\textbf{Subset} & \multicolumn{3}{c}{\adjustbox{max width=1.26cm}{\strut \textbf{CBraMod}}} & \multicolumn{3}{c}{\adjustbox{max width=1.26cm}{\strut \textbf{CodeBrain}}} & \multicolumn{3}{c}{\adjustbox{max width=1.26cm}{\strut \textbf{CSBrain}}} & \multicolumn{3}{c}{\adjustbox{max width=1.26cm}{\strut \textbf{EEGPT}}} & \multicolumn{3}{c}{\adjustbox{max width=1.26cm}{\strut \textbf{NeuroRVQ}}} & \multicolumn{3}{c}{\adjustbox{max width=1.26cm}{\strut \textbf{REVE-base}}} & \multicolumn{3}{c}{\adjustbox{max width=1.26cm}{\strut \textbf{ST-EEGFormer-small}}} & \multicolumn{3}{c}{\adjustbox{max width=1.26cm}{\strut \textbf{LEAD}}} & \multicolumn{3}{c}{\adjustbox{max width=1.26cm}{\strut \textbf{HEAR-base}}} & \multicolumn{3}{c}{\adjustbox{max width=1.26cm}{\strut \textbf{HEAR-large}}}\\
\cmidrule(lr){2-4}\cmidrule(lr){5-7}\cmidrule(lr){8-10}\cmidrule(lr){11-13}\cmidrule(lr){14-16}\cmidrule(lr){17-19}\cmidrule(lr){20-22}\cmidrule(lr){23-25}\cmidrule(lr){26-28}\cmidrule(lr){29-31}
 & B-ACC & Kap. & F1 & B-ACC & Kap. & F1 & B-ACC & Kap. & F1 & B-ACC & Kap. & F1 & B-ACC & Kap. & F1 & B-ACC & Kap. & F1 & B-ACC & Kap. & F1 & B-ACC & Kap. & F1 & B-ACC & Kap. & F1 & B-ACC & Kap. & F1\\
\end{longtable}
\endgroup

\FloatBarrier

\FloatBarrier



\end{document}